\documentclass[final,3p,times]{elsarticle}

\usepackage{amssymb}
\usepackage{amsmath}
\usepackage{indentfirst}
\usepackage{graphicx}
\usepackage{color}
\usepackage{url}
\usepackage{bigstrut}
\usepackage{multirow,booktabs}
\usepackage{algorithm}
\usepackage{algorithmic}
\usepackage{subfigure}
\newcommand{\tabincell}[2]{\begin{tabular}{@{}#1@{}}#2\end{tabular}}
\journal{Information Sciences}

\begin{document}

\begin{frontmatter}
\title{A Disease Diagnosis and Treatment Recommendation System Based on Big Data Mining and Cloud Computing}

\author[Addr1]{Jianguo~Chen}
\ead{jianguochen@hnu.edu.cn}
\author[Addr1,Addr2]{Kenli Li\corref{cor1}}
\ead{lkl@hnu.edu.cn}
\author[Addr1]{Huigui~Rong} 
\ead{ronghg@hnu.edu.cn}
\author[Addr4]{Kashif~Bilal}
\ead{kashif@qu.edu.qa}
\author[Addr6]{Nan~Yang}
\ead{Shmily.1989.2008@stu.xjtu.edu.cn}
\author[Addr1,Addr3]{Keqin Li}
\ead{lik@newpaltz.edu}

\cortext[cor1]{Corresponding author. Tel.: +86-731-88664161.}

\address[Addr1]{College of Computer Science and Electronic Engineering, Hunan University, Changsha, Hunan 410082, China}
\address[Addr2]{National Supercomputing Center in Changsha, Changsha, Hunan 410082, China}
\address[Addr4]{Comsats Institute of Information Technology, Abbottabad 45550, Pakistan}
\address[Addr6]{The second affiliated hospital of Xi'an Jiaotong University, Xi'an, Shaanxi 710049, China}
\address[Addr3]{Department of Computer Science, State University of New York, New Paltz, NY 12561, USA}

\begin{abstract}
It is crucial to provide compatible treatment schemes for a disease according to various symptoms at different stages.
However, most classification methods might be ineffective in accurately classifying a disease that holds the characteristics of multiple treatment stages, various symptoms, and multi-pathogenesis.
Moreover, there are limited exchanges and cooperative actions in disease diagnoses and treatments between different departments and hospitals.
Thus, when new diseases occur with atypical symptoms, inexperienced doctors might have difficulty in identifying them promptly and accurately.
Therefore, to maximize the utilization of the advanced medical technology of developed hospitals and the rich medical knowledge of experienced doctors, a Disease Diagnosis and Treatment Recommendation System (DDTRS) is proposed in this paper.
First, to effectively identify disease symptoms more accurately, a Density-Peaked Clustering Analysis (DPCA) algorithm is introduced for disease-symptom clustering.
In addition, association analyses on Disease-Diagnosis (D-D) rules and Disease-Treatment (D-T) rules are conducted by the Apriori algorithm separately.
The appropriate diagnosis and treatment schemes are recommended for patients and inexperienced doctors, even if they are in a limited therapeutic environment.
Moreover, to reach the goals of high performance and low latency response, we implement a parallel solution for DDTRS using the Apache Spark cloud platform.
Extensive experimental results demonstrate that the proposed DDTRS realizes disease-symptom clustering effectively and derives disease treatment recommendations intelligently and accurately.
\end{abstract}


\end{frontmatter}

\section{Introduction}
\label{intro}
\subsection{Motivation}
Technological advancements and cost reduction in medical equipment and disease diagnosis have greatly accelerated the adoption of state-of-the-art technologies in various hospitals \cite{cc029, cc030}.
The benefits of obtaining interactive and intelligent medical service based on knowledge discovery are rapidly growing.
The accurate classification of different disease symptoms is essential in helping doctors carry out compatible treatment schemes for the disease.
In contrast, traditional disease classification methods usually follow naive practices based on limited disease information, which might fail to further classify a disease according to symptoms at different treatment stages.
In particular, for diseases with the characteristics of multiple similar treatment stages, various symptoms, and multi-pathogenesis, the accuracy and effectiveness of traditional classification algorithms are significantly lower.
Therefore, it is crucial to find suitable approaches to accurately classify disease symptoms based on inspection reports.

In general, medical doctors diagnose diseases and select treatment schemes based mostly on their personal experience and knowledge.
Inadequate communication, experience exchange, and cooperation between young and senior doctors results in the failure of young doctors to learn and take guidance from the experience, diagnoses, and treatment plans of experienced senior doctors.
For example, the determinants of patients' and doctors' delays in the diagnosis and treatment of colorectal cancer were discussed in \cite{cc035}.
Despite the generation and availability of abundant medical data related to patients, diseases, treatment plans and their results, these data are not appropriately analyzed to extract useful knowledge and not efficiently shared among doctors and hospitals.
Due to the lack of diagnosis experience, fledgling medical doctors might have difficulty in correctly diagnosing a disease in a patient that has an atypical symptom; thus, they are clueless in prescribing effective treatment plans.
Hence, the sharing and recommendation of medical knowledge can help fledgling doctors elevate their disease diagnosis and treatment experience.
A recommendation system of disease diagnosis and treatment is developed to find a balance between the medical resources of developed and underdeveloped hospitals and between the medical knowledge of experienced doctors and inexperienced doctors.

Because of the massive volume, variety, and continuous updating of medical data, the efficient processing of medical data and the real-time response of the treatment recommendation has become an important issue.
Fortunately, parallel computing and cloud computing provide powerful capabilities to cope with large-scale data.
Various hospitals have developed cloud-computing-based solutions for treatment guidance and have implemented various improvement measures for medical services.
For example, Abbas \emph{et al}. discussed a cloud-based health insurance plan recommendation system that implements a user-centered approach \cite{cc001}.
Apache Hadoop \cite{cc002} is a famous cloud platform that is widely utilized in big data mining.
Li \emph{et al}. proposed an efficient tool for de novo peptide sequencing that utilizes the Hadoop cloud computing environment \cite{cc019}.
Apache Spark \cite{cc003} is an excellent cloud platform that is suitable for data mining with iterative computation.
Parallel programming models of Resilient Distributed Datasets (RDDs) and Directed Acyclic Graphs (DAGs) are supported by Spark, which is built in a memory computing framework.
Benefitting from the RDD and DAG models, data caches are saved in memory, and iterations are performed on the same dataset directly from memory.
Hence, the Spark platform is more suitable for data mining with iterative computation by saving huge amounts of disk I/O operation time.

\subsection{Our Contributions}
In this paper, we focus on medical resource sharing and treatment intelligence in big data and cloud computing environments and propose a Disease Diagnosis and Treatment Recommendation System (DDTRS).
Large-scale historical inspection datasets are analyzed to derive disease-symptom clusters.
Association relationships between diseases, diagnoses, and treatments are discovered in historical treatment records.
Based on these relationships, valuable diagnosis and treatment plans of diseases are recommended to medical doctors and patients according to the current disease stages.
Extensive experimental analysis indicates that the system recommends solutions effectively and accurately.
Our contributions in this paper are summarized as follows.

\begin{itemize}
  \item To effectively identify more accurate disease symptoms, especially for diseases with multiple treatment stages and multi-pathogenesis, a Density-Peak-based Clustering Analysis (DPCA) algorithm is introduced for patients' disease-symptom clustering, depending on the symptoms extracted from large-scale historical inspection data.
  \item To profusely utilize and share the valuable disease diagnosis and treatment knowledge of experienced doctors and developed hospitals, Disease-Diagnosis (D-D) and Disease-Treatment (D-T) association rules are defined and analyzed by the Apriori algorithm.
  \item Interactive recommendation interfaces of DDTRS are designed and implemented for medical doctors and patients.
         Medical doctors and patients can access the inspection reports and the corresponding treatment recommendations at different treatment stages.
         They can update the inspection results in DDTRS to obtain recommendations.
  \item To achieve the goals of high performance and low latency response, we parallelize DDTRS on the Apache Spark cloud computing platform.
      Massive volumes of medical data are stored in the Hadoop Distributed File System (HDFS), and a parallel solution is employed based on the RDD programming model.
\end{itemize}

The remainder of the paper is organized as follows.
Section 2 reviews the related work.
Section 3 introduces a disease diagnosis and treatment recommendation system, which consists of three core modules: a DPCA-based disease-symptom clustering process, a disease-treatment association analysis process, and recommendation interfaces.
To reach the goals of efficiency and timeliness, the proposed system is parallelized in Section 4 using the Apache Spark cloud computing platform.
Experimental and application results are presented in Section 5 with respect to recommendation effectiveness and performance.
Finally, Section 6 concludes the paper with a discussion of future work and research directions.

\section{Related Work}
Benefitting from the development of medical technology and information technology, numerous studies focus on the fields of disease prevention, disease treatment, hospital informatization, and drug discovery.
Applications of big data analytics in hospitals were adopted in \cite{cc025,cc005}.
Patients' healthcare datasets were stored digitally in the form of Electronic Health Records (EHRs), and realistic and valuable information was obtained from these records by using felicitous analysis techniques and software tools.
Diverse patterns, trends, associations, visualization, querying, information privacy, and predictive analytics of the healthcare datasets were analyzed.
Evidence-Based Medicine (EBM) is a medical method that establishes best-practice recommendations based on graded treatment schemes for diagnostic and therapeutic issues in health care \cite{cc018}.
In EBM, decisions about the care of individual patients are made based on the current best available clinical evidence, combined with the doctor's clinical experience and taking into account the patient's values and aspirations.
As a valuable and applicable medical method, EBM has been utilized in various specialties, such as neurology \cite{cc027}, pediatric urology \cite{cc034}, and burn care \cite{cc025}.
In addition, numerous efforts have been made to use big data and data mining technology for EBM \cite{cc041, cc030}.
For example, in \cite{cc041}, Yesha \emph{et al}. introduced a personalized decision support system to enhance EBM using big data analytics.
Reports about medication accidents and treatment failures due to diagnosis and treatment delays or inappropriate diagnoses were reviewed in \cite{cc029}.
Makwakwa \emph{et al}. considered health system delays in the cases of diagnosis, new treatment and retreatment of pulmonary tuberculosis in Malawi \cite{cc022}.
The median patient delay was 14.0 days for both new treatment and retreatment of Tuberculosis cases, and the median health system delay was 59.0 days for new treatment and 40.5 days for retreatment cases.
The authors concluded that effective management and new diagnostic techniques were needed for both new treatment and retreatment cases.
In our previous work, we explored a parallel patient-treatment-time prediction algorithm and its applications in hospital queuing recommendations in big data environments \cite{cc005}.
There are limited research and applications in disease diagnosis and treatment recommendation based on large-scale historical medical data.

With a focused on clustering analysis, abundant notable achievements were contributed in existing studies \cite{cc012, cc015}.
Multifarious traditional clustering algorithms were reviewed in \cite{cc042}, which consist of partitioning methods, hierarchical methods, density-based methods, grid-based methods, and model-based methods.
Distinct from these traditional clustering approaches, groups of novel density-peak-based clustering algorithms were explored \cite{cc037, cc038, cc039}.
A density-peak-based hierarchical clustering method (DenPEHC) was proposed, in which different clusters could be generated directly at each possible clustering layer.
Alex \emph{et al}. addressed the DPCA algorithm based on the idea that a cluster center is surrounded by neighbors that have low local density and large delta distance from any point that has a higher density \cite{cc031}.
A fast density-based data stream clustering algorithm for mixed data was presented in \cite{cc006} with self-determined cluster centers.
Other valuable clustering algorithms were introduced in \cite{cc014, cc032}.
Compared with iterative clustering algorithms, the computational complexity of DPCA is very low.
Benefitting from the robustness on rough data and the non-iterative training of DPCA, we introduce the DPCA algorithm to cluster disease symptoms in this work.

Efforts are made in the fields of association analysis and information recommendation \cite{cc043,cc045,cc016}.
In \cite{cc043}, Yuan \emph{et al}. discussed a personalized recommendation algorithm based on user models and a user-project matrix.
A random-walk-based recommendation algorithm was suggested in \cite{cc045} that considers item categories.
Jia \emph{et al}. improved the Apriori algorithm based on association analysis \cite{cc016}.
In \cite{cc033}, Sing \emph{et al}. reviewed automatic tag recommendation algorithms for social recommender systems.
Corbellini \emph{et al}. \cite{cc008} explored an architecture and platform for developing distributed recommendation algorithms, in which the recommendation time is reduced by modifying job distribution and non-invasive tuning strategies.
In the medical field, valuable patient-oriented recommendations were introduced in \cite{cc046, cc007, cc017}.
Zheng \emph{et al}. introduced a context neighbor recommender platform, which integrates contexts via neighbors for recommendations \cite{cc046}.
Chen \emph {et al}. proposed an autocratic decision-making system using group recommendation methods \cite{cc007}.
In \cite{cc017}, Kefalas \emph{et al}. explored a graph-based taxonomy approach for recommendation algorithms and systems.
A wearable assistant for gait training for Parkinson's disease was introduced in \cite{cc024}, which works with the freezing of gait in out-of-the-laboratory environments.
In addition, large-scale medical data were analyzed in \cite{cc009} to identify frequent diseases using the Apriori algorithm.
The frequencies of diseases were identified for patients living in various geographical locations during different periods.
Taking advantage of the high accuracy of the Apriori algorithm, we exploit it to extract associated rules of disease diagnosis and treatment schemes.

The rapid development of big data and cloud computing technologies provides powerful computing ability  for mining large-scale medical data.
Valuable knowledge can be extracted from large-scale data in each application field.
Yao \emph{et al}. designed and developed a big medical data processing system \cite{cc040}.
Shunmei \emph{et al}. reviewed recommendation methods for big data applications based on cloud computing \cite{cc026}.
Moreover, cloud computing technology provides powerful computing power for machine learning and big data mining.
Numerous machine-learning and data-mining algorithms were implemented on various cloud platforms \cite{cc010, cc047, cc036}, such as Apache Hadoop and Apache Spark.
Li \emph{et al}. introduced a computing resource management framework of cloud computing platforms, which can be applied in the field of medical data mining \cite{cc020}.
In \cite{cc010}, Sara \emph{et al}. used the MapReduce model of Apache Hadoop to enhance the performance of the Random Forest (RF) algorithm for imbalanced big data.
In \cite{cc036}, Wang \emph{et al}. developed a new Distributed Trajectory R-Tree (DTR-Tree) index algorithm on the Apache Spark platform and achieved high efficiency.
Heuristically, we implement the parallel solution of DDTRS on the Apache Spark cloud computing platform.

\section{Disease Diagnosis and Treatment Recommendation System}
\label{section3}
To effectively identify more accurate disease symptoms and profusely utilize and share the rich medical knowledge of experienced doctors, we propose a disease diagnosis and treatment recommendation system.
DDTRS is comprised of two core modules:
(a) the disease-symptom clustering analysis module, in which a DPCA-based clustering algorithm is introduced to classify disease symptoms based on the given inspection reports,
and (b) the disease diagnosis and treatment recommendation module.
Effective association rules of disease diagnosis and treatment are defined and analyzed by the Apriori algorithm.
Valuable diagnosis and treatment plans are recommended for medical doctors and patients via the interactive interfaces of DDTRS.

\begin{figure}[!ht]
\setlength{\abovecaptionskip}{0pt}
\setlength{\belowcaptionskip}{0pt}
\centering
\includegraphics[width=3.0in]{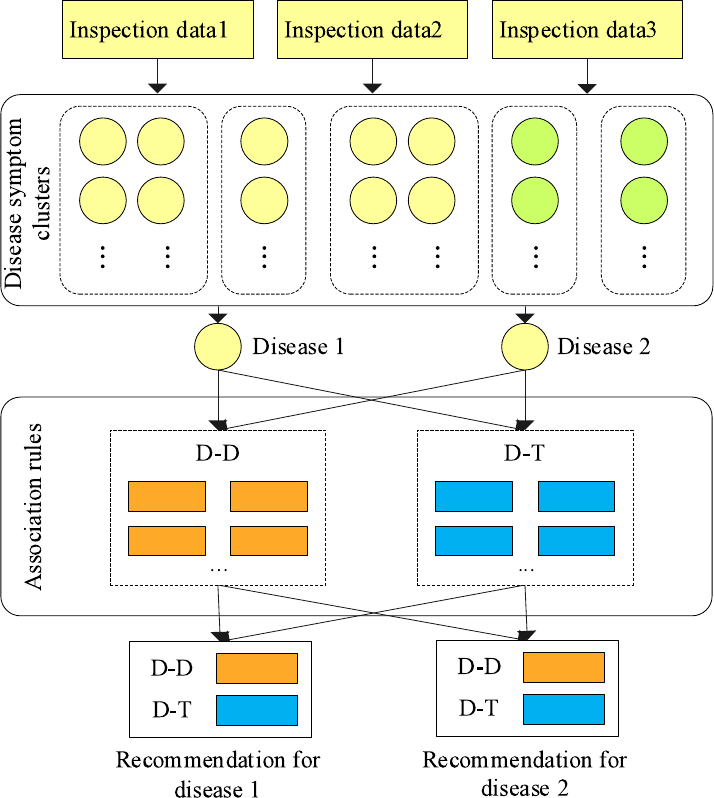}
\caption{Workflow of disease diagnosis and treatment recommendation}
\label{fig01}
\end{figure}

In the early stages of a patient's treatment, the patient usually undergoes a series of inspection tests, as advised by the attending doctor.
Once the inspection reports are available, they are submitted to DDTRS to obtain a disease-symptom cluster, and then further obtain recommendations of disease diagnosis and treatment plans.
The workflow of the disease diagnosis and treatment recommendation is presented in Figure \ref{fig01}.

\subsection{Standardization Process of Medical Data}
Currently, for some inspection tasks, different standards are used in different hospitals, and the standards of one hospital might not be acceptable to another.
Few hospitals use unified standards for inspection reports under the management of superior health administrative departments.
Therefore, a standardization process is required for the medical datasets gathered from different hospitals.
Large-scale historical medical datasets are gathered from our cooperating hospitals.
Private patient information, such as ID code, patient name, telephone, and address, is filtered before the data collection process.
In the standardization process, a series of operations are performed for the inspection datasets before performing the clustering and association analysis, such as data integration and data cleaning.

\subsubsection{Preprocessing of Patient Inspection Data}
Profiting from precise instruments and quantifiable data, inspection datasets are utilized as the data source of disease-symptom clustering.
In the preprocessing phase, inspection datasets from different departments and different hospitals are collected.
Due to the diverse contents of the inspection data, the inspection data are in multiple data formats, e.g., numeric, text, and image formats.

\textbf{(1) Inspection data in numeric format.}

Some of the inspection-task-generated report datasets are stored in numeric format, such as those for routine blood inspections, routine urine inspections, and bone marrow examinations.
There might be multiple inspection items, which should be tested in each inspection task.
An example report of a routine blood inspection task is shown in Table \ref{table01}.

\begin{table}[!ht]
\centering
\caption{Example of a blood inspection report}
\label{table01}
\tabcolsep1pt
\begin{tabular}{clcccc}
\hline
No. & ~~~~~~~~Inspection items & Result & Unit & Reference value range & Remark  \\
\hline
1 & White blood cell (WBC)  &  5.86 &	$10^{9}$/L	& 4-10    & \\
2 & Red blood cell (RBC)	&  5.6	&   $10^{12}$/L	& 3.5-5.5 & $\uparrow$ \\
3 & Platelet (PLT)	        &  341	&   $10^{9}$/L	& 100-300 & $\uparrow$ \\
4 & Hematocrit (HCT)	    &  47.3	&   \%	        & 37-52   & \\
5 & Hemoglobin (HGB)	    &  148.0&	g/L  	    & 110-170 & \\
6 & Mean corpuscular hemoglobin (MCH)	& 28.3	& pg	& 27-35	& \\
7 & Mean corpuscular hemoglobin concentration (MCHC) & 314	& g/L	& 320-362	& $\downarrow$ \\
8 & Mean corpuscular volume (MCV)	    & 85.9	& fL	& 82.6-99.1	&\\
9 & Mean platelet volume (MPV)	        & 11.5	& fL	& 7.6-13.2 &\\
10& Platelet larger cell ratio (P-LCR)	& 28.5	& \%	& 13-43 & \\	
  & ...                     & ...   & ...           &... & \\
\hline
\end{tabular}
\end{table}

As shown in Table \ref{table01}, more than 10 inspection items are required in a blood inspection task.
The patient information and the corresponding inspection items of each inspection task are selected as the feature variables of the patient's disease data.
Due to differences in inspection items, the numbers of feature variables of the disease data are different among inspection tasks.
The feature variables of the inspection datasets selected for disease-symptom clustering analysis are presented in Table \ref{table02}.
Detailed descriptions of the inspection item names and abbreviations are given in Table \ref{table03}.

\begin{table}[!ht]
\setlength{\abovecaptionskip}{0pt}
\setlength{\belowcaptionskip}{0pt}
\renewcommand{\arraystretch}{1.3}
\caption{Feature variables of the inspection data (partial lists)}
\centering
\label{table02}
\tabcolsep1pt
\begin{tabular}{cll}
\hline
No.&	Inspection tasks&	~~~~~~~~~~~~~~~~~~~~~~~Feature variables \\
\hline
1 & Routine blood & \tabincell{l}{ specimen, patient gender, age, WBC, RBC, PLT, HCT, HGB, MCH, MCHC, MCV,\\ MPV, P-LCR, ...} \\
2 & Routine urine &  \tabincell{l}{specimen, patient gender, age, BLD, BIL, URO, KET, PRO, NIT, GLU, PH, SG,\\ WBC, RBC, ...} \\
3 & Tumor marker &  \tabincell{l}{specimen, patient gender, age, AFP, AFU, LN, CEA, FER, CA50, CA199, CA125,\\ CA153, TNF-$\alpha$, NSE, ...} \\
4 & Thyroid function &  \tabincell{l}{specimen, patient gender, age, ALT, AST, TP, ALB, GLO, TBIL, DBIL, IBIL,\\ ALP, CHE, ADA, MAO, LDH, ...} \\
5 & Hepatitis B surface antigen &  \tabincell{l}{specimen, patient gender, age, ALT, ALP, GGT, TP, ALB, TBA, HBsAb,\\ MAO, HBeAb, HBcAb, AFU, ...} \\
 & ... & ... \\
\hline
\end{tabular}
\end{table}

\begin{table}[!ht]
\caption{Inspection item names and abbreviations}
\centering
\label{table03}
\tabcolsep1pt
\begin{tabular}{cclccl}
\hline
No.& Abbreviation&	Full name of inspection items & No.& Abbreviation&	Full name of inspection items\\
\hline
1 & ADA & Adenosine deaminase           & 27 & HCT & Hematocrit \\
2 & AFP & Alphafetoprotein              & 28 & HGB & Hemoglobin \\
3 & AFU & $\alpha$-fucosidase           & 29 & IBIL & Indirect bilirubin\\
4 & ALB & Albumen                       & 30 & KET & Urine acetone bodies (Ketone) \\
5 & ALP & Alkaline phosphatase          & 31 & LDH & Lactate dehydrogenase\\
6 & ALT & Alanine aminotransferase      & 32 & LN  & Laminin\\
7 & AST & Aspartate aminotransferase    & 33 & MAO & Monamine oxidase\\
8 & BIL & Bilirubin                     & 34 & MCH & Mean corpuscular hemoglobin\\
9 & BLD & Urine occult blood            & 35 & MCHC & Mean corpuscular hemoglobin concentration\\
10 & CA50 & Carbohydrate antigen 50     & 36 & MCV & Mean corpuscular volume  \\
11 & CA125 & Carbohydrate antigen 125   & 37 & MPV & Mean platelet volume\\
12 & CA153 & Carbohydrate antigen 153   & 38 & NIT & Nitrite\\
13 & CA199 & Carbohydrate antigen 199   & 39 & NSE & Neuro-specific enolase\\
14 & CEA & Carcinoembryonic antigen     & 40 & RBC & Red blood cell \\
15 & CHE & Choline esterase             & 41 & PH & pH value\\
16 & DBIL & Direct bilirubin            & 42 & P-LCR & Platelet larger cell ratio\\
17 & EC & Epithelial cells              & 43 & PLT & Platelet \\
18 & FER & Ferritin                     & 44 & PRO & Urine protein\\
19 & GGT & Gamma glutamyl transpeptidase   & 45 & SG & Specific gravity of urine\\
20 & GLO & Globulin                     & 46 & TBA & Total bile acid\\
21 & GLU & Urine glucose                & 47 & TBIL & Total bilirubin\\
22 & GOL & Urine color                  & 48 & TNF-$\alpha$ & Tumour necrosis factor-$\alpha$\\
23 & HBcAb & Hepatitis B core antibody  & 49 & TP & Total protein\\
24 & HBeAb & Hepatitis B e antibody     & 50 & URO & Urobilinogen \\
25 & HBsAb & Hepatitis B surface antibody   & 51 & WBC & White blood cell \\
26 & HBsAg & Hepatitis B surface antigen    & 52 & $\beta$-HCG & $\beta$-human chorionic gonadotophin\\
... & ... & ...   & ... & ... & ... \\
\hline
\end{tabular}
\end{table}

\textbf{(2) Inspection data in text format.}

Distinct from the inspection tasks addressed above, there exists another type of inspection tasks, namely, imaging tasks, e.g., Computer Tomography (CT) scanning, Magnetic Resonance Imaging (MRI) scanning,  X-ray, type-B sonography, and color doppler ultrasound.
The symptom descriptions of these inspection tasks are expressed in the form of images or texts.
The patient information and the corresponding symptom results of each inspection task are chosen as the feature variables of patient inspection data.

\subsubsection{Preprocessing of Disease Diagnosis and Treatment Data}

\textbf{(1) Disease diagnosis-scheme data.}

In this work, the diagnosis-scheme data refer to the detailed diagnosis descriptions of diseases recorded by doctors rather than the simple definitions of the diseases.
Diverse diagnosis schemes are provided at different treatment stages of the same disease.
These schemes are recorded by medical doctors or assistants in text format.

$DSCs = \{DSC_{1}, ~DSC_{2}, ~..., ~DSC_{i}, ~..., ~DSC_{n}\}$ is assumed to be a dataset of disease-symptom clusters.
The diagnosis schemes of the same disease-symptom cluster are collected for association analysis.
Let $DSs = \{DS_{1}, ~DS_{2}, ~...,$ $~DS_{j}, ~..., ~DS_{m}\}$ be a dataset of diagnosis schemes.
The dataset of the association items between the disease-symptom clusters and the corresponding diagnosis schemes is denoted as $E(DSC, ~DS)$.
An association item $e_{ij} = (DSC_{i}, ~DS_{j})$ is defined when there exists a diagnosis scheme $DS_{j}$ for disease-symptom cluster $DSC_{i}$, and $e_{ij} \in E(DSC, ~DS)$.

\textbf{(2) Disease treatment-scheme data.}

Available treatment schemes in hospitals include injections, intravenous infusion{s}, surgical treatments, needle therapy, cupping therapy, and physical therapy.
In the treatment process of a disease, the medical doctor might adjust the treatment plans for the patient depending on the current inspection results.
Taking a disease as an example, treatment scheme $A$ is suggested for a serious condition, $B$ for a moderate condition, and $C$ for a mild condition.
Namely, a specific treatment scheme or a treatment-scheme combination is performed for the current treatment stage of a disease.

Let $TSs = \{TS_{1}, ~TS_{2}, ~..., ~TS_{j}, ~..., ~TS_{m}\}$ be a dataset of treatment schemes.
The dataset of the associated items between the disease-symptom clusters and the corresponding treatment schemes is denoted as $E(DSC, ~TS)$.
An associated item $e_{ij} = (DSC_{i}, ~TS_{j})$ is defined when there exists a treatment scheme $TS_{j}$ for disease-symptom cluster $DSC_{i}$, and $e_{ij} \in E(DSC, ~TS)$.

\subsection{DPCA-based Disease-Symptom Clustering}
\label{section_DPCA}
Although the traditional classification measures are widely used in disease classification and have achieved notable results, they might have difficulty in further classifying a disease accurately according to the symptoms at different treatment stages.
To identify disease symptoms more accurately and effectively, the DPCA algorithm is introduced for patients' disease-symptom clustering.
The DPCA algorithm is appropriate for diseases that have the characteristics of multiple similar treatment stages, various symptoms, and multi-pathogenesis.
Taking advantage of the non-iterative training and the robustness of DPCA on rough data, we introduce DPCA to cluster the disease symptoms in this section.
DPCA is an innovative density-peak-based clustering algorithm, which is based on the idea that a cluster center is surrounded by neighbors that have low local densities and large delta distances.
After determining the cluster centers, each of the remaining data points is classified to the nearest neighboring cluster that has a higher density.

\subsubsection {Clustering Analysis on Inspection Data in Numeric Format}

\begin{figure}[!ht]
\setlength{\abovecaptionskip}{0pt}
\setlength{\belowcaptionskip}{0pt}
\centering
\includegraphics[width=4.0in]{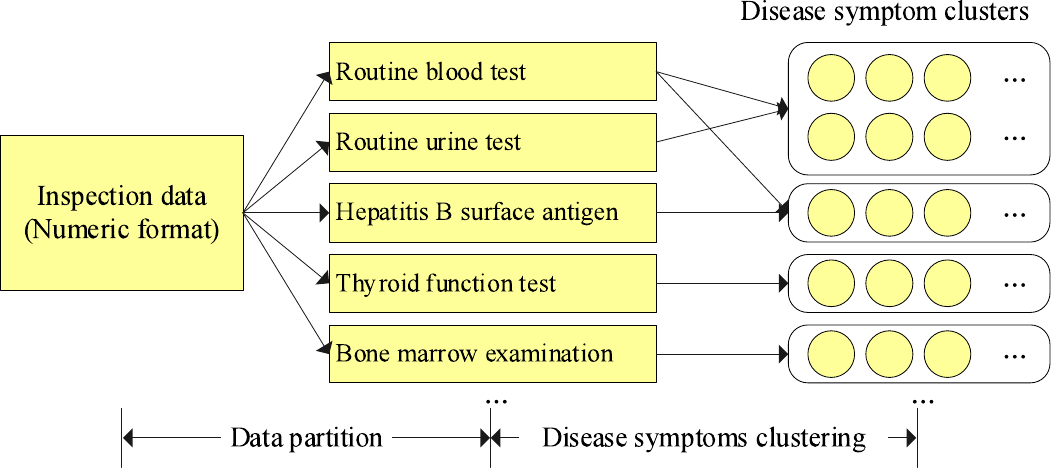}
\caption{Workflow of clustering analysis on the inspection data}
\label{fig02}
\end{figure}

Because of the different dimensions of the datasets in separate inspection tasks, the pre-processed inspection datasets are grouped by task name.
The dataset of each inspection task is clustered by DPCA to obtain disease-symptom clusters separately.
The workflow of the clustering analysis on the inspection data in numeric format is illustrated in Figure \ref{fig02}.

Let $X$ be a dataset of the pre-processed inspection data in numeric format, defined as follows:

\begin{center}
$X = \{X_{1}, ~X_{2}, ~..., ~X_{k}, ~..., ~X_{m}\}$,
\end{center}
where $m$ is the number of inspection tasks and $X_{k}$ is the data subset of the $k$-th inspection task.
Due to the different dimensions of these data subsets, each data subset $X_{k}$ is clustered by the DPCA algorithm separately.
The steps of the disease-symptom clustering are described as follows.

\textbf{(1) Calculate the local density for each data point.}

Assume that inspection data subset $X_{k}$ contains $M$ feature variables, namely, the dimension of $X_{k}$ is $M$.
There are $N$ inspection records in $X_{k}$, namely, $|X_{k}| = N$.
We calculate the distance matrix $D_{k} = (d_{ij})_{N \times N}$ of $X_{k}$, where $d_{ij}$ denotes the distance between $x_{i}$ and $x_{j}$, as defined in Eq. (\ref{eq01}):

\begin{equation}
\label{eq01}
d_{ij} =\sqrt{\sum_{a=1}^{M}{(x_{ia} - x_{ja})^{2}}},
\end{equation}
where $x_{ia}$ and $x_{ja}$ are the $a$-th feature variables of $x_{i}$ and $x_{j}$, respectively.
In the original DPCA algorithm, for each data point $x_{i}$ in $X_{k}$, the local density $\rho_{i}$ is calculated based on $d_{ij}$ between different data points.
The local density $\rho_{i}$ of $x_{i}$ is defined in Eq. (\ref{eq02}):

\begin{equation}
\label{eq02}
\rho_{i} = \sum_{j}{\chi (d_{ij} - d_{c})},
\end{equation}
where $d_{c}$ is a cutoff distance and $\chi (x) = 1$ if $(d_{ij} - d_{c}) < 0$; otherwise, $\chi (x) = 0$.

Numerous shortcomings can be observed in Eq. (\ref{eq02}), such as low robustness, discrete distribution, and inaccurate density-peak points.
$\rho_{i}$ is equal to the number of data points for which the distance from $x_{i}$ is smaller than $d_{c}$.
Therefore, the cutoff distance $d_{c}$ greatly influences the local density $\rho$, leading to low robustness of the algorithm.
In addition, depending on the calculation method of $\chi (\cdot)$ in Eq. (\ref{eq02}), the local density $\rho_{i}$ is an integer value, and the local density values of all data points in $X_{k}$ obey a discrete distribution.
Based on the discrete values of local density, it is difficult to detect local density-peak points and cluster centers accurately.
Moreover, in cases of multi-density and non-uniform distribution, there might exist multiple inaccurate density-peak points in each cluster.

To address the limitations of Eq. (\ref{eq02}) and effectively cluster the datasets with multi-density and non-uniform distribution, we improve the DPCA algorithm by proposing an optimization solution for the calculation of the local density.
A Radial Basis Function (RBF) kernel function is introduced to measure the local density.
The RBF kernel function $K(||x - x_{c}||)$ is a monotonic function of the Euclidean distance between any data point $x$ and a data center $x_{c}$ in a space.
The Gauss Kernel Function (GKF), defined in Eq. (\ref{eq03}), is a widely used RBF kernel function:

\begin{equation}
\label{eq03}
K(||x-x_{c}||) = e^{- \frac{||x-x_{c}||^{2}}{2 \times \sigma^{2}}},
\end{equation}
where $||x-x_{c}||$ is the center of the kernel function and $\sigma$ is a width parameter, which controls the radial range of the function.
$\sigma$ is set as $d_{c}$ in this work.
The calculation method of the local density $\rho_{i}$ is optimized based on GKF, namely, Eq. (\ref{eq02}) can be modified to Eq. (\ref{eq04}):

\begin{equation}
\label{eq04}
\begin{aligned}
\rho_{i} &= \sum{K(||x-x_{c}||)}, \\
         &= \sum{e^{- \frac{||x-x_{c}||^{2}}{(2\sigma)^{2}}}},\\
         &= \sum_{j \in N, j \neq i}{e^{- \left( \frac{d_{ij}}{2d_{c}} \right)^{2}}}.
\end{aligned}
\end{equation}

In Eq. (\ref{eq02}), the density of a data point is obtained by counting its neighbors.
In contrast, in Eq. (\ref{eq04}), the density of a data point is obtained by calculating a monotonic function of the distances of all the other data points to the current data point, which can accurately reflect the density distribution of the data points in the entire dataset.
GKF maps the data sample to a higher-dimensional space, and it can cope with a nonlinear relationship between the class labels and data features.
Owing to the continuous distribution of density values, it can accurately determine the densities of different data points and thereby identify the density peaks.
Compared with Eq. (\ref{eq02}), the local density $\rho$ based on GKF in Eq. (\ref{eq04}) is minimally impacted by the cutoff distance, and the improved DPCA algorithm achieves higher robustness.
Benefitting from the advantages of GKF, even for a dataset under a uniform density distribution, the local density peak points can be obtained accurately.

Based on the local density $\rho_{i}$, the delta distance $\delta_{i}$ of each point $x_{i}$ is calculated.
$\delta_{i}$ is obtained by computing the minimum distance between $x_{i}$ and the data points that have higher density.
The delta distance $\delta_{i}$ of data point $x_{i}$ is defined in Eq. (\ref{eq05}):

\begin{equation}
\label{eq05}
\delta_{i} =
\left\{
\begin{array}{ll}
\max_{j}{d_{ij}} & \rho_{i} = max(\rho),\\
\min_{j:\rho_{j} < \rho_{i}}{d_{ij}} & otherwise.
\end{array}
\right.
\end{equation}
If the data point $x_{i}$ has the highest density, then $\delta_{i} = \max_{j}{d_{ij}}$.

\textbf{(2) Identify cluster centers based on $\rho$ and $\delta$.}

\begin{figure}[!ht]
\setlength{\abovecaptionskip}{0pt}
\setlength{\belowcaptionskip}{0pt}
\centering
\includegraphics[width=3.0in]{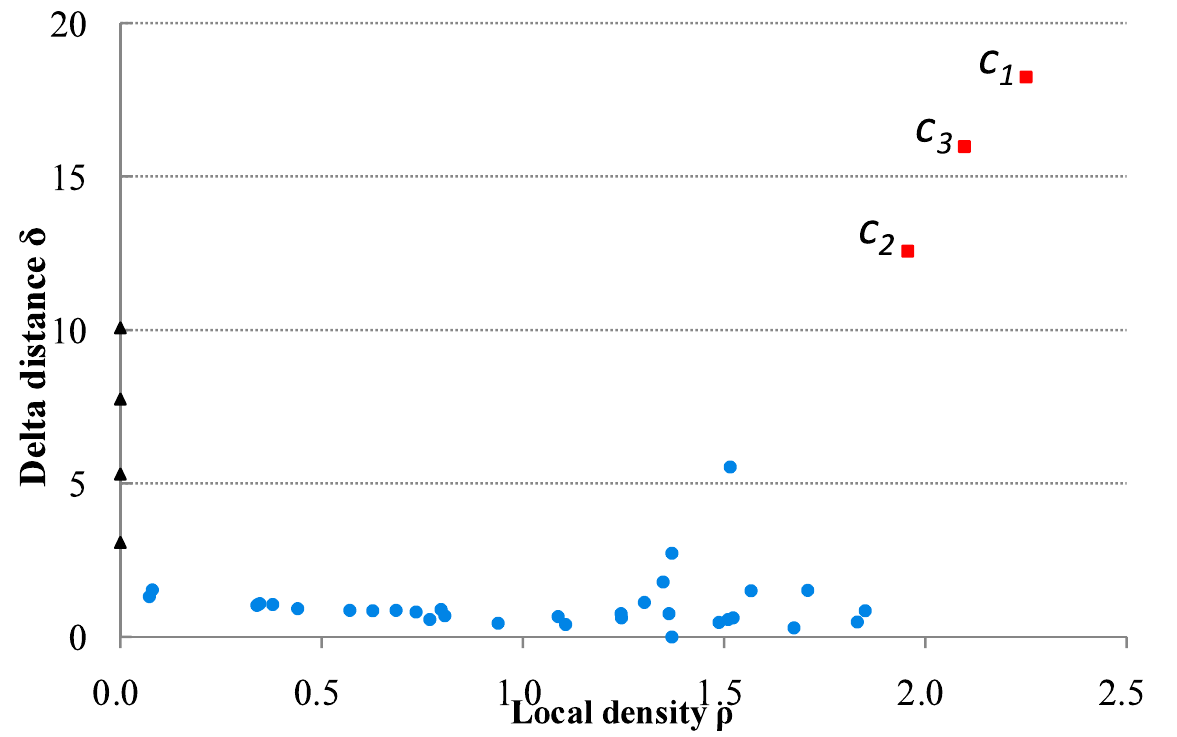}
\caption{Decision graph for the DPCA-based disease-symptom clustering}
\label{fig03}
\end{figure}

On the basis of $\rho$ and $\delta$, disease-symptom cluster centers are identified.
Data points that have relatively high values of $\rho$ and $\delta$ are designated cluster centers.
Data points that have small densities $\rho$ and high delta distances $\delta$ are considered outliers.
A decision graph of $\rho$ and $\delta$ is drawn for the disease-symptom clusters detection.
An example of a decision graph for disease-symptom clustering is shown in Figure \ref{fig03}.

In Figure \ref{fig03}, there are three decision points with higher values of both local density $\rho$ and delta distance $\delta$, which are shown as red squares in the figure.
These decision points are identified as candidate disease-symptom cluster centers of the current inspection data subsets.
In contrast, the decision points indicated by black triangles have low values of local density and high values of delta distance.
These points are identified as outliers and removed from the clustering results.
The decision  points shown as blue circles in the figure are the remaining data points, which are neither the density peaks nor the noise data.
These remaining data points will be assigned to the related disease-symptom clusters in the next step.

\textbf{(3) Assign the remaining data points to the nearest disease-symptom clusters.}

\begin{figure}[!ht]
\setlength{\abovecaptionskip}{0pt}
\setlength{\belowcaptionskip}{0pt}
\centering
\includegraphics[width=4.0in]{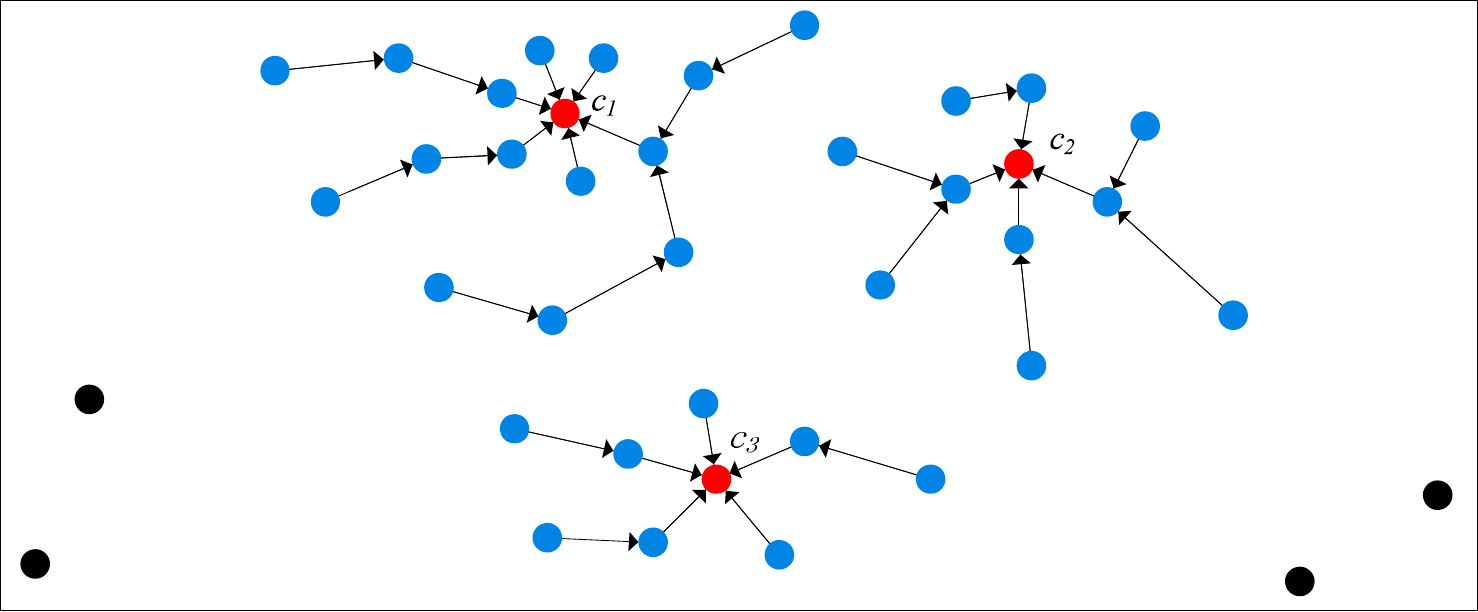}
\caption{Example of the assignment of the remaining data points}
\label{fig04}
\end{figure}

After the disease-symptom cluster centers are detected, each of the remaining data points is assigned to the cluster to which the nearest and higher-density neighbors belong.
For each remaining data point $x_{i}$, the set of neighbors with higher density is denoted as $N^{'}(x_{i})$.
The data point $x_{j} \in N^{'}(x_{i})$ with the minimum distance $d_{ij}$ is discovered from distance matrix $D_{k}$.
If $x_{j}$ has been assigned to a cluster $c_{a}$, then $x_{i}$ is also assigned to $c_{a}$.
Otherwise, the cluster of $x_{j}$ is further determined iteratively.
This step is repeated until all the remaining data points are assigned to the clusters.
The DPCA-based clustering analysis process of the inspection data in numeric format is completed and disease-symptom clusters are obtained.
An example of the assignment of the remaining data points is shown in Figure \ref{fig04}.
The process of the DPCA-based disease-symptom clustering analysis of the inspection data in numeric format is presented in Algorithm \ref{alg1}.

\begin{algorithm}[!ht]
\caption{DPCA-based disease-symptom clustering analysis on the inspection data in numeric format}
\label{alg1}
\begin{algorithmic}[1]
\REQUIRE ~\\
    $X$: the inspection data in numeric format;\\
    $P_{d_{c}}$: the percentage of distance for cutoff distance; \\
    $\varepsilon_{\rho}$: threshold value of the local density for cluster-center decision;\\
    $\varepsilon_{\delta}$: threshold value of the delta distance for cluster-center decision.\\
\ENSURE ~\\
    $DSC$: the disease-symptom clusters.
\FOR {each inspection task $X_{k}$ in $X$}
    \STATE $D$ $\leftarrow$ calculate distance $d_{ij}$ from $X_{k}$;
    \STATE $D^{'}$ $\leftarrow$ arrange $D$ in ascending order;
    \STATE $d_{c}$ $\leftarrow$ distance of the $P_{d_{c}}$ position in $D^{'}$;
    \FOR {each $x_{i}$ in $X_{k}$}
       \STATE $\rho_{i}$ $\leftarrow$ calculate local density $\sum_{j \in N, j \neq i}{e^{- \left(\frac{d_{ij}}{2d_{c}} \right)^{2}}}$;
       \STATE $\delta_{i}$ $\leftarrow$ calculate delta distance $\min_{j:\rho_{j} < \rho_{i}}{d_{ij}}$;
       \IF {($\rho_{i}$ bigger than $\varepsilon_{\rho}$ and $\delta_{i}$ bigger than $\varepsilon_{\delta}$)}
           \STATE ${DSC_{c}}$ $\leftarrow$ identify cluster centers ($\rho_{i}, ~\delta_{i}$);
       \ENDIF
    \ENDFOR
    \FOR {each $x_{i}$ in $X_{k}$}
        \STATE append $x_{i}$ to the nearest cluster $c_{a}$;
    \ENDFOR
\ENDFOR
\RETURN $DSC$.
\end{algorithmic}
\end{algorithm}

In DPCA, the cluster assignment process is executed in one step.
The time complexity of Algorithm \ref{alg1} is $O(M {\overline{n}}^{2})$, where $M$ is the number of inspection tasks and $\overline{n}$ is the average number of records in dataset $X_{k}$ of each inspection task.
Thus, the space complexity of the DPCA algorithm is $O( {\overline{n}}^{2})$.
Compared with iterative clustering algorithms, the computational complexity of DPCA is very low.

\subsubsection {Clustering Analysis on Inspection Data in Text Format}

\begin{figure}[!ht]
\setlength{\abovecaptionskip}{0pt}
\setlength{\belowcaptionskip}{0pt}
\centering
\includegraphics[width=5.4in]{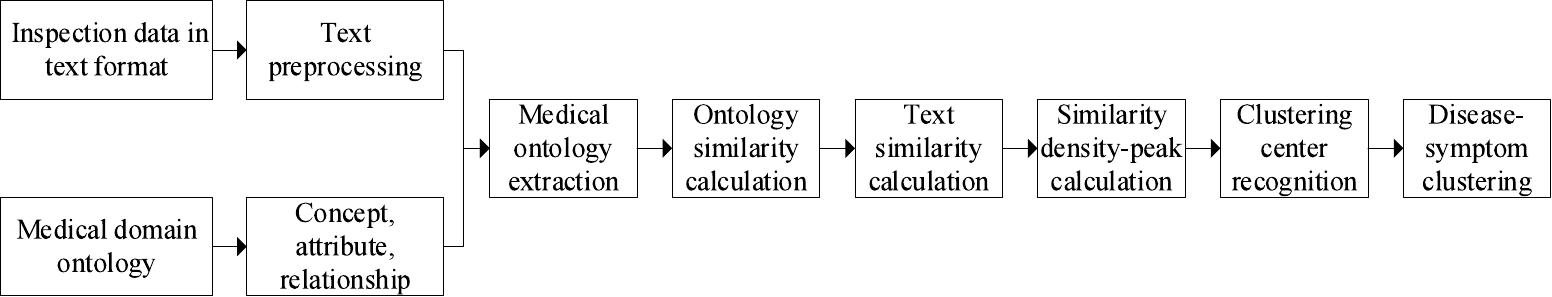}
\caption{Workflow of clustering analysis on the inspection data in text format}
\label{fig05}
\end{figure}

In the disease diagnosis process, in addition to quantitative inspection tasks, admitting doctors record the conditions of patients' diseases by observing and consulting the patients' situations.
These records of disease diagnostic information are saved in text format and constitute another important basis for judging the characteristics of patients' diseases.
Considering the specific terminology and rigor of the medical domain, we introduce a text clustering analysis algorithm for inspection data in text format based on a medical ontology model in this section.
First, inspection data in text format are collected and preprocessed for medical word segmentation, and a medical domain ontology library is constructed.
Second, medical ontologies of the pre-processed inspection data are extracted based on the medical domain ontology library.
In addition, the ontological similarity of these medical ontologies and the text similarity of these textual samples are calculated.
Finally, on these bases, the DPCA algorithm is applied on these text data to obtain the corresponding disease symptom clusters, in which the local density peak $\rho$ is calculated based on the ontological and textual similarities.
The workflow of the clustering analysis on the inspection data in text format is shown in Figure \ref{fig05}.

\textbf{(1) Construct medical domain ontology library.}

Before the ontology extraction, the inspection dataset in text format is preprocessed, in which the private information of patients and doctors is filtered.
Then, specific words of the inspection text are segmented using the Natural Language Processing and Information Retrieval (NLPIR) segmentation system \cite{cc044}.
In NLPIR, the text contents are segmented by using the $N$-shortest-path rough cut method based on the dictionary with word frequency and property statistical information to obtain the best rough coverage results.
In addition, properties of segmented words are marked automatically after obtaining the overall optimal word segmentation results.

\begin{figure}[!ht]
\setlength{\abovecaptionskip}{0pt}
\setlength{\belowcaptionskip}{0pt}
\centering
\includegraphics[width=4.0in]{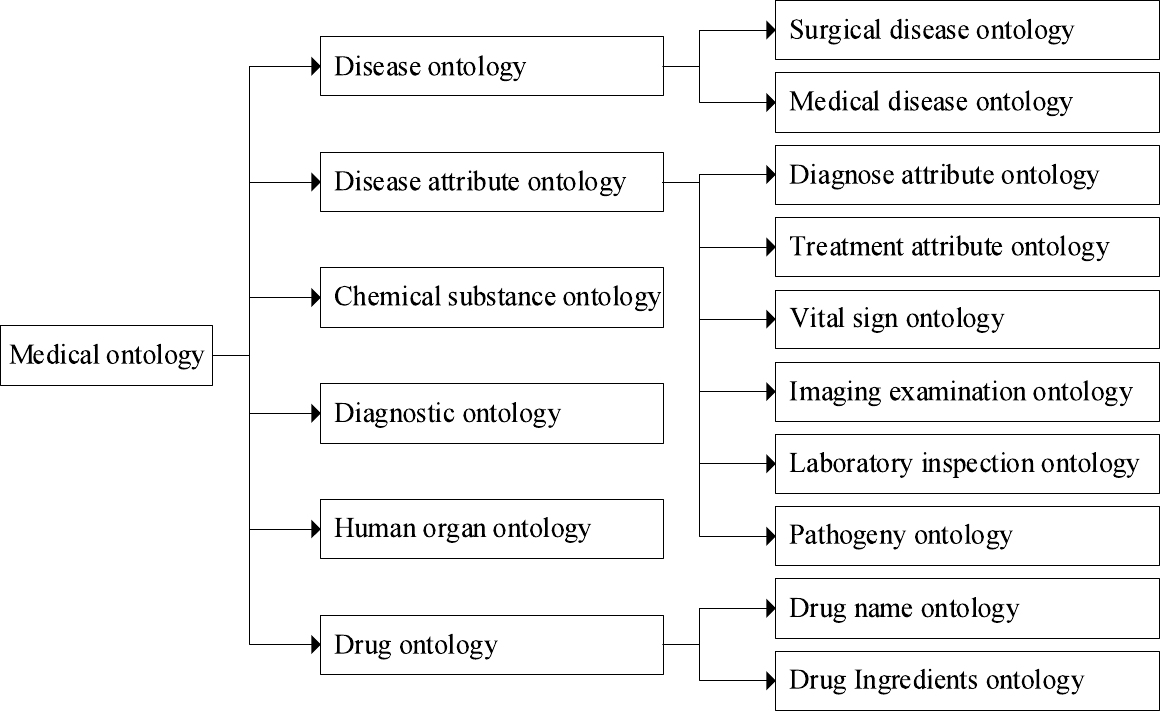}
\caption{Architecture of the medical domain ontology library (partial)}
\label{fig06}
\end{figure}

An ontology is a formal expression of the relationship between a set of concepts and their relationships in a specific field.
As a method of expressing knowledge, ontologies have been widely used in various application fields, among which biomedicine is one of the most active areas of ontology applications \cite{cc021, cc023}.
PROTEGE \cite{cc004} is a free and open-source ontology editor and a knowledge-based framework developed by the Stanford BioMedical Informatics Research Center.
Rim \emph{et al}. introduced a medical domain ontology construction approach for a medical decision-support system \cite{cc011}.
In \cite{cc021}, Lu \emph{et al}. proposed a medical ontology-enhanced text processing method for infectious disease informatics.
In this work, we construct a medical domain ontology library for the text clustering analysis of the inspection data in text format.
The medical domain ontology library is built based on the data from PubMed \cite{cc028}, which is a database that provides free searching of biomedical papers and abstracts.
We extracted numerous useful phrases and words from the related medical literature in the PubMed biomedical database.
The noun phrases are extracted and marked as the concepts of biomedical applications, and the verb phrases are transformed into the corresponding relationships among these concepts.
An example of the architecture of the medical domain ontology library in this work is shown in Figure \ref{fig06}.

\textbf{(2) Extract the medical ontology of the text inspection dataset.}

Based on the medical domain ontology library, we extract various medical ontology objects from the preprocessed text inspection datasets.
Because the ontology consists of concepts and relationships, the medical concepts and relationships are extracted separately.
Let $O = \{C, ~R\}$ be the medical ontology set in this work, where $C$ is the set of medical concepts and $R$ is the set of relationships.
The medical concepts of the text inspection dataset are composed of $k$ semantic elements, as defined in Eq. (\ref{eq06}):

\begin{equation}
\label{eq06}
C = \left\{ \bigcup_{i=1}^{k} c_{i} \right\},
\end{equation}
where $c_{i}$ refers the $i$-th semantic element in the medical domain.
Medical ontology relationships consist of medical concepts and their relationships, namely, $R_{i} = \{ c_{i1}, ~c_{i2}, r_{i} \}$.

\textbf{(3) Calculate the similarity of medical ontologies and text samples.}

We introduce an ontology-based similarity measure method to calculate the similarity of medical ontologies and text samples.
As described above, the medical ontology is composed of medical concepts and relationships.
Hence, the ontological similarity metrics include concept similarity and relationship similarity measures.
For two ontology objects $o_{i}$ and $o_{j}$, the concept similarity between them is defined in Eq. (\ref{eq07}):

\begin{equation}
\label{eq07}
S_{c}(o_{i},~o_{j}) = \frac{1} {n_{i} \times n_{j}}\sum_{i \leq n_{i}, j \leq n_{j}}{sim(c_{i}, ~c_{j})},
\end{equation}
where $n_{i}$ and $n_{j}$ are the numbers of medical concepts of $o_{i}$ and $o_{j}$, respectively.
If $c_{i} = c_{j}$, then $sim(c_{i}, ~c_{j})=1$; otherwise, $sim(c_{i}, ~c_{j})=0$.
The relationship similarity of ontology objects $o_{i}$ and $o_{j}$ is defined in Eq. (\ref{eq08}):

\begin{equation}
\label{eq08}
\begin{aligned}
S_{r}(o_{i},~o_{j}) &= \frac{1} {m_{i} \times m_{j}}\sum_{i \leq m_{i}, j \leq m_{j}}{sim(R_{i}, ~R_{j})}, \\
                    &= \frac{1} {m_{i} \times m_{j}}\sum_{i \leq m_{i}, j \leq m_{j}}{sim(c_{i1}, ~c_{j1}) \times sim(c_{i2}, ~c_{j2}) \times sim(r_{i}, ~r_{j})},
\end{aligned}
\end{equation}
where $m_{i}$ and $m_{j}$ are the numbers of medical ontology relationships of $o_{i}$ and $o_{j}$, respectively.
Based on the concept and relationship similarity measure methods, the similarity of two ontology objects is defined in Eq. (\ref{eq09}):

\begin{equation}
\label{eq09}
S (o_{i},~o_{j}) = S_{c}(o_{i},~o_{j}) \times S_{r}(o_{i},~o_{j}).
\end{equation}

Each inspection record in text format is composed of a series of ontologies.
Assume that there are two inspection text sample $x_{a} = \{o_{a1}, ~..., o_{a\alpha}\}$ and $x_{b} = \{o_{b1}, ~..., o_{db\beta}\}$.
The similarity of $x_{a}$ and $x_{b}$ is defined in Eq. (\ref{eq10}):

\begin{equation}
\label{eq10}
S (x_{a},~x_{b}) = \frac{1}{\alpha \times \beta} \sum_{i \leq \alpha, j \leq \beta}{S(o_{ai},~o_{bj})},
\end{equation}
where $\alpha$ and $\beta$ are the numbers of medical ontologies of text samples $x_{a}$ and $x_{b}$, respectively.

\textbf{(4) Perform text clustering analysis of text inspection data based on PDCA algorithm.}

Based on the medical ontological similarity, we analyze the disease symptoms from text inspection data by using the PDCA algorithm.
The main steps in clustering analysis of the inspection data in text format are same as those used for the inspection data in numeric format.
We calculate the local density $\rho$ and delta distance $\delta$ for each data sample.
Unlike for the inspection data in numeric format, the distance between a pair of data points for the inspection data in text format is measured by the similarity of the documents instead of the Euclidean metric.
Therefore, the local density $\rho_{i}$ of data sample $x_{i}$ based on GKF can be calculated by Eq. (\ref{eq11}):

\begin{equation}
\label{eq11}
\begin{aligned}
\rho_{i} &= \sum{K(||x-x_{c}||)}, \\
         &= \sum{e^{- \frac{||x-x_{c}||^{2}}{(2\sigma)^{2}}}},\\
         &= \sum_{j \in N, j \neq i}{e^{- \left( \frac{S_{ij}}{2S_{c}} \right)^{2}}},
\end{aligned}
\end{equation}
where $S_{ij} = S (x_{i},~x_{j})$ and $S_{c}$ is a cutoff similarity.
Afterwards, the delta distance $\delta$ of each data sample $x_{i}$ is calculated subsequently with Eq. (\ref{eq12}):
\begin{equation}
\label{eq12}
\delta_{i} =
\left\{
\begin{array}{ll}
\max_{j}{S_{ij}} & \rho_{i} = max(\rho),\\
\min_{j:\rho_{j} < \rho_{i}}{S_{ij}} & otherwise.
\end{array}
\right.
\end{equation}
If data sample $x_{i}$ has the highest density, then $\delta_{i} = \max_{j}{S_{ij}}$.
We identify cluster centers based on $\rho$ and $\delta$.
Data points that have relatively higher values of $\rho$ and $\delta$ are designated cluster centers, while data points that have a small value of $\rho$ and a high value of $\delta$ are considered outliers.
Afterwards, the remaining data samples are assigned to the disease-symptom cluster to which the nearest higher-density neighbors belong.
The clustering analysis process of the inspection data in text format based on the medical domain ontology and the DPCA algorithm is completed and disease-symptom clusters are obtained.

\subsection{Disease Diagnosis and Treatment Recommendation}

Based on the disease-symptom clustering, the multifarious human intelligence of medical doctors is accumulated in the form of valuable diagnosis and treatment knowledge.
We perform association analysis of the D-D and D-T rules.
Moreover, appropriate disease treatment plans are recommended to medical doctors and patients via DDTRS.

\subsubsection{Association Analysis of Disease Diagnosis and Treatment Schemes}
\begin{figure}[!ht]
\setlength{\abovecaptionskip}{0pt}
\setlength{\belowcaptionskip}{0pt}
\centering
\includegraphics[width=0.75\columnwidth]{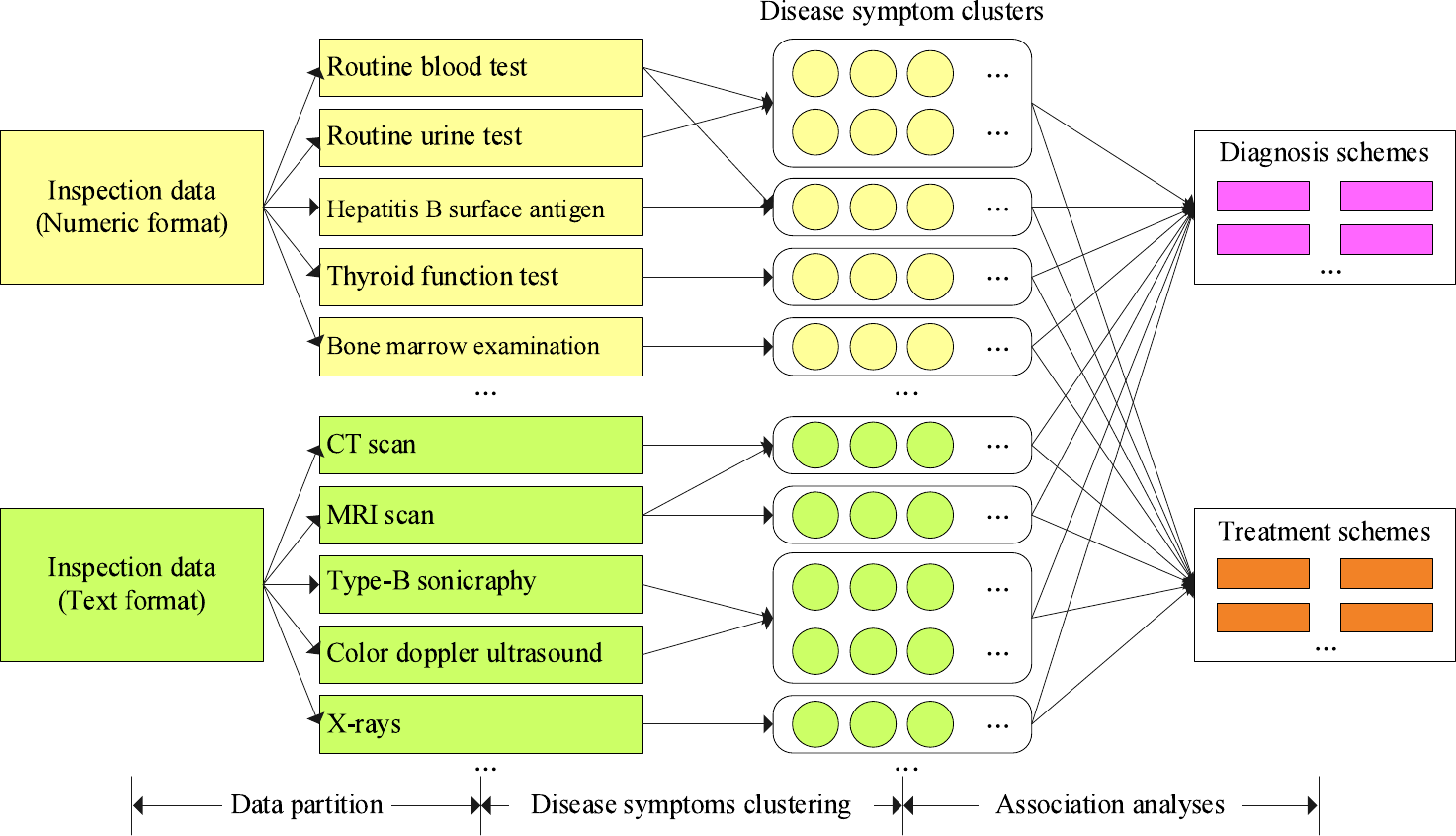}
\caption{Workflow of disease diagnosis- and treatment-scheme association analysis}
\label{fig07}
\end{figure}

After preprocessing the disease diagnosis- and treatment-scheme data, the Apriori algorithm is introduced for the association analysis of the D-D rules and the D-T rules, respectively.
The Apriori algorithm is used to detect the association relationships between the disease-symptom clusters and the disease diagnosis and treatment schemes that have been utilized successfully.
The workflow of the disease diagnosis- and treatment-scheme association analysis is illustrated in Figure \ref{fig07}.
Owing to the similar principles of three rules, we take the D-T rules as an example to elaborate the association analysis process.
The detailed steps of the Apriori-based disease treatment-scheme association analysis are described as follows.

In general, for a patient, multiple visits are required in the treatment process of a disease.
The disease diagnosis and treatment data collected during each patient visit are considered as an association record in this work.
Each association record contains the inspection reports, diagnosis schemes, and corresponding treatment schemes.
These datasets are collected and grouped by disease into the same clusters, which are utilized as the data source of association analysis.

\textbf{(1) Set minimum $Support$ and minimum $Confidence$.}

To obtain the association rules between each disease and its treatment schemes, the treatment schemes of the same disease-symptom cluster are analyzed separately.
The D-T rules of disease-symptom cluster $DSC_{i}$ are defined as $R_{ix} \Rightarrow R_{iy}$, where $R_{ix} \subset E(DSC, ~TS)$, $R_{iy} \subset E(DSC, ~TS)$, and $R_{ix} \bigcap R_{iy} = \emptyset $.

\textbf{Definition 1: (Support).} \textit{The support of an association rule $R_{ix} \Rightarrow R_{iy}$ in the association rule set $E(DSC, ~TS)$ refers to the ratio of the number of treatment records that contain both $R_{ix}$ and $R_{iy}$ to the total number of treatment records for disease-symptom cluster $DSC_{i}$.}

The $Support$ of association rule $R_{ix} \Rightarrow R_{iy}$ is denoted as $Sup(R_{ix} \Rightarrow R_{iy})$, as defined in Eq. (\ref{eq13}):

\begin{equation}
\label{eq13}
Sup(R_{ix} \Rightarrow R_{iy}) =\frac{|R_{ix} \bigcup R_{iy} \subset E(DS, ~TS)|}{|E(DS, ~TS)|}.
\end{equation}

\textbf{Definition 2: (Confidence).} \textit{The confidence of an association rule $R_{ix} \Rightarrow R_{iy}$ in the association relationship set $E(DSC, ~TS)$ refers to the ratio of the number of treatment records that contain both $R_{ix}$ and $R_{iy}$ to the number of association rules that contain $R_{ix}$.}

The $Confidence$ of association rule $R_{ix} \Rightarrow R_{iy}$ is denoted as $Conf(R_{ix} \Rightarrow R_{iy})$, as defined in Eq. (\ref{eq14}):

\begin{equation}
\label{eq14}
\begin{aligned}
Conf(R_{ix} \Rightarrow R_{iy}) &=\frac{Sup(R_{ix} \Rightarrow R_{iy})}{Sup(R_{ix})},\\
                             &=\frac{|R_{ix} \bigcup R_{iy} \subset E(DS, ~TS)|}{| R_{ix} \subset E(DS, ~TS)|}.
\end{aligned}
\end{equation}

\textbf{Definition 3: (Frequent Itemsets).} \textit{If the support and confidence of an association rule $R_{ix} \Rightarrow R_{iy}$ in $E(DSC, ~TS)$ are greater than the minimum support degree and the minimum confidence, respectively, the rule is defined as a frequent item $FI \in E(DS, ~TS)$.}

For a dataset $TSs$ of treatment schemes, the aim of association rule mining is to extract the D-T rules that satisfy the minimum $Support$ and minimum $Confidence$ at the same time.
According to the results of disease-symptom clustering analysis and treatment schemes, the minimum support is set as $MinSup(R_{ix} \Rightarrow R_{iy}) = 2$ and the minimum $Confidence$ is defined as $MinConf (R_{ix} \Rightarrow R_{iy}) = 50\%$.

\textbf{(2) Generate Frequent Itemsets.}

The association rule itemset $FI(DSC, ~TS) = \{R_{ij}\}$ is generated as the frequent itemsets of D-T rules, where $Sup(R_{ix} \Rightarrow R_{iy}) > MinSup(R_{ix} \Rightarrow R_{iy})$ and $Conf(R_{ix} \Rightarrow R_{iy}) > MinConf (R_{ix} \Rightarrow R_{iy})$.

\textbf{(3) Extract strong association rules from frequent itemsets.}

After obtaining the maximum frequent itemsets of the treatment-scheme data, candidate D-T rules are extracted from these frequent itemsets.
These D-T rules are utilized in the disease treatment recommendation.
The Apriori-based association analysis of disease diagnosis and treatment scheme is presented in Algorithm \ref{alg3}.

\begin{algorithm}[!ht]
\caption{Apriori-based association analysis of diseases' diagnosis and treatment scheme}
\label{alg3}
\begin{algorithmic}[1]
\REQUIRE ~\\
    $DSCs$: the dataset of disease-symptom clusters;\\
    $TSs$: the dataset of disease treatment-scheme records;\\
    $MinSup$: the predefined value for the minimum $Support$;\\
    $MinConf$: the predefined value for the minimum $Confidence$.\\
\ENSURE ~\\
    $FI(DSC, ~TS)$: the frequent itemsets of the association rules.
\STATE create frequent itemset $FI(DSC, ~TS)$;
\FOR {each $DSC_{i}$ in $DSCs$}
    \FOR {each $TS_{j}$ in $DSC_{i}.TSs$}
        \STATE $R_{ix}$ $\leftarrow$ get treatment association $E(DSC_{i}, ~TS_{j})$;
        \STATE $R_{iy}$ $\leftarrow$ get treatment association $E(DSC_{i}, ~TS_{j+1})$;
        \STATE $Sup(R_{ix} \Rightarrow R_{iy})$ $\leftarrow$ $\frac{|R_{ix} \bigcup R_{iy} \subset E(DSC, ~TS)|}{|E(DSC, ~TS)|}$;
        \IF {$Sup(R_{ix} \Rightarrow R_{iy}) \geq MinSup$}
            \STATE $Conf(R_{ix} \Rightarrow R_{iy})$ $\leftarrow$ $\frac{Sup(R_{ix} \Rightarrow R_{iy})}{Sup(R_{ix})};$
            \IF {$Conf(R_{ix} \Rightarrow R_{iy}) \geq MinConf$}
                \STATE $FI(DSC, ~TS)$ $\leftarrow$ append frequent itemset $\{R_{ix}, ~R_{iy}\}$;
            \ENDIF
        \ENDIF
    \ENDFOR
\ENDFOR
\RETURN $FI(DSC, ~TS)$.
\end{algorithmic}
\end{algorithm}

Similar to the case of the D-T rules, the association analysis of the D-D rules is carried out using the Apriori algorithm.
Thus, strong association rules among the disease-symptom clusters, diagnosis schemes, and treatment schemes are extracted.

\subsubsection{Recommendation Interfaces}

On the basis of the disease diagnosis- and treatment-scheme association analysis, we implement a Client / Service (C/S) application for medical doctors and a mobile application for patients to provide treatment recommendations.
Using the applications' interactive interfaces, medical doctors and patients can access the inspection reports and the corresponding treatment recommendations at different treatment stages.
They can update the inspection results in the system to obtain recommendations.
DDTRS can adjust the output according to the input changes.

\textbf{(1) Interactive recommendation interfaces for doctors.}

\begin{figure}[!ht]
\setlength{\abovecaptionskip}{0pt}
\setlength{\belowcaptionskip}{0pt}
\centering
\includegraphics[width=0.9\columnwidth]{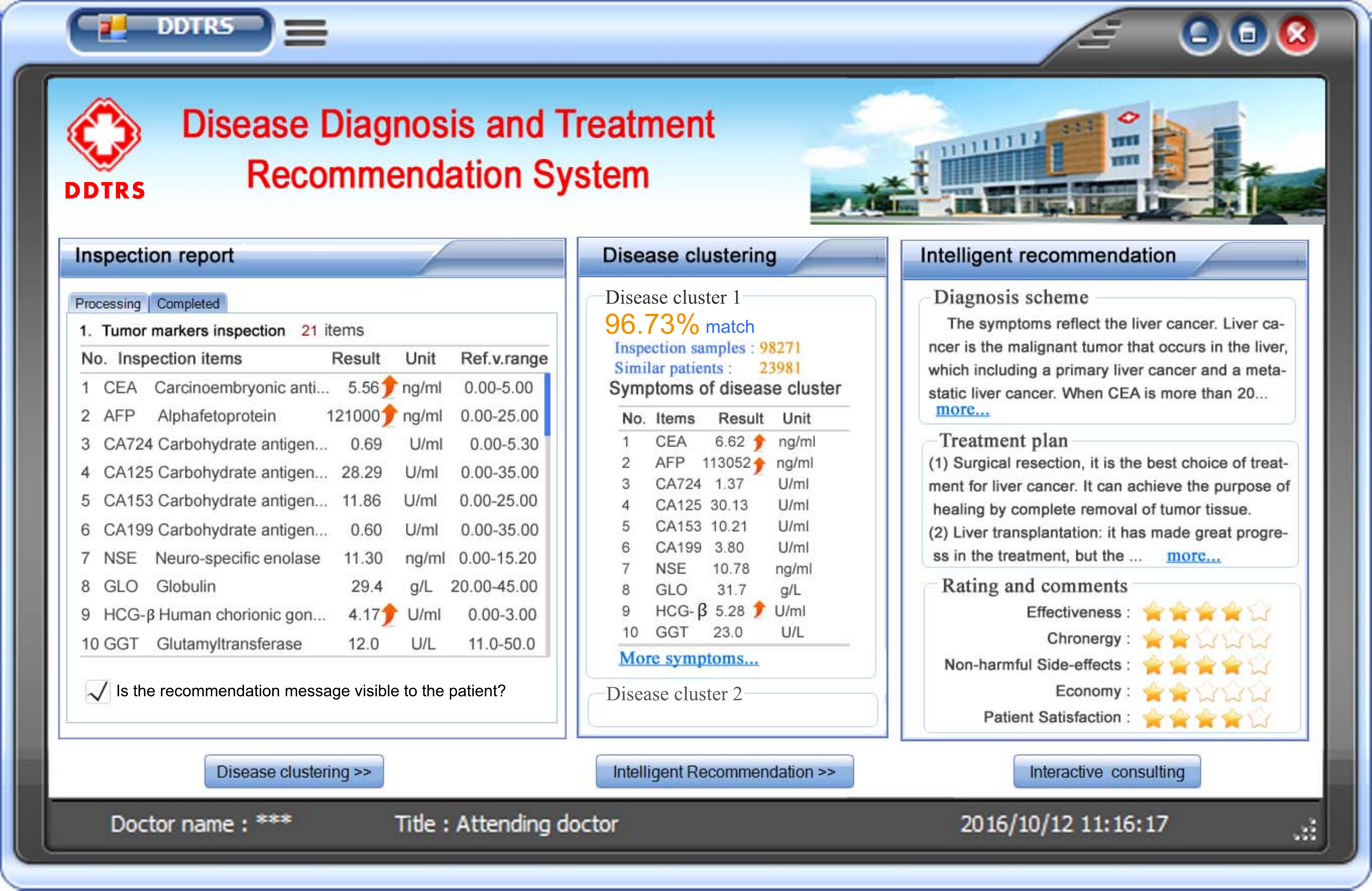}
\caption{Recommendation interface of DDTRS for medical doctors}
\label{fig08}
\end{figure}

We provide interactive recommendation interfaces of DDTRS to medical doctors in hospitals.
An example of one such interface is shown in Figure \ref{fig08}.
To facilitate used by local patients, the application is designed using the Chinese language.
However, the application can be easily customized to different languages.
For the convenience of the readers, here we illustrate the English version of the application.
From the interactive interface, medical doctors can review the inspection reports of their patients and submit the inspection data to the disease-symptom clustering module.
Based on the results of disease-symptom clustering, the diagnosis scheme and treatment plans for each disease are recommended to the medical doctors.

In Figure \ref{fig08}, the inspection report shows that abnormal results were obtained for multiple inspection items, such as CEA, AFP, and HCG-$\beta$.
Among them, CEA reaches 5.56 $ng/ml$, which is slightly higher than the normal reference range of 0.00 - 5.00;
AFP reaches 121000 $ng/ml$, which is significantly higher than the normal reference range of 0.00 - 25.00;
and HCG-$\beta$ reaches 4.17 $U/ml$, which is higher than the normal reference range of 0.00 - 3.00.
After clustering the submitted inspection report, a disease-symptom cluster is obtained from the similar historical inspection data. The corresponding items of the cluster center are listed as follows: CEA = 6.62 $ng/ml$, AFP = 113052 $ng/ml$, and HCG-$\beta$ = 5.28 $U/ml$.
The cluster is termed Liver Cancer, which is detected in the form of mean symptom values.
Depending on the disease-symptom cluster Liver Cancer, related diagnosis schemes and treatment plans are recommended, as shown in the right region of Figure \ref{fig08}.
Further description of the case analysis of liver cancer is presented in Section 5.3.1.

From a medical point of view, inappropriate treatment recommendations might lead to misdiagnosis and endanger the patient's health.
Therefore, the correctness and effectiveness of the recommendations are the key issues for this kind of application.
The quality of a treatment scheme for a patient can be evaluated by tracking the changes of the items' values in his inspection reports.
However, due to different pathologies and diverse patient health conditions, it is difficult to implement a common evaluation standard for the quality of each treatment scheme.
Doctors might provide feedback if some incorrect recommendations are made.
A feedback collection and analysis module should be utilized to improve the effectiveness of the application.
Evaluation is performed based on the quality indicators of the treatment recommendations, which are fed back from medical doctors.

When a doctor receives a recommended treatment scheme, he can adopt and apply it completely or in part to his patient.
After tracking and comparing the changes of the inspection reports and patient's health condition, the doctor determines whether the recommended treatment scheme is effective for the patient's disease.
Afterwards, via the application interface, he submits the quality indicators of the treatment scheme, such as effectiveness, chronergy, non-harmful side-effects, economy, and patient satisfaction.
The value of each indicator is in the range of (1 $\sim$ 5).
The detailed description of the quality-evaluation method of treatment schemes is presented in Section \ref{section_QualityEvaluation}.

\textbf{(2) Interactive recommendation interfaces for patients.}

\begin{figure*}[!ht]
\setlength{\abovecaptionskip}{0pt}
\setlength{\belowcaptionskip}{0pt}
\centering
 \subfigure[Inspection report]{
 \label{fig09:a}
 \includegraphics[width=1.4in]{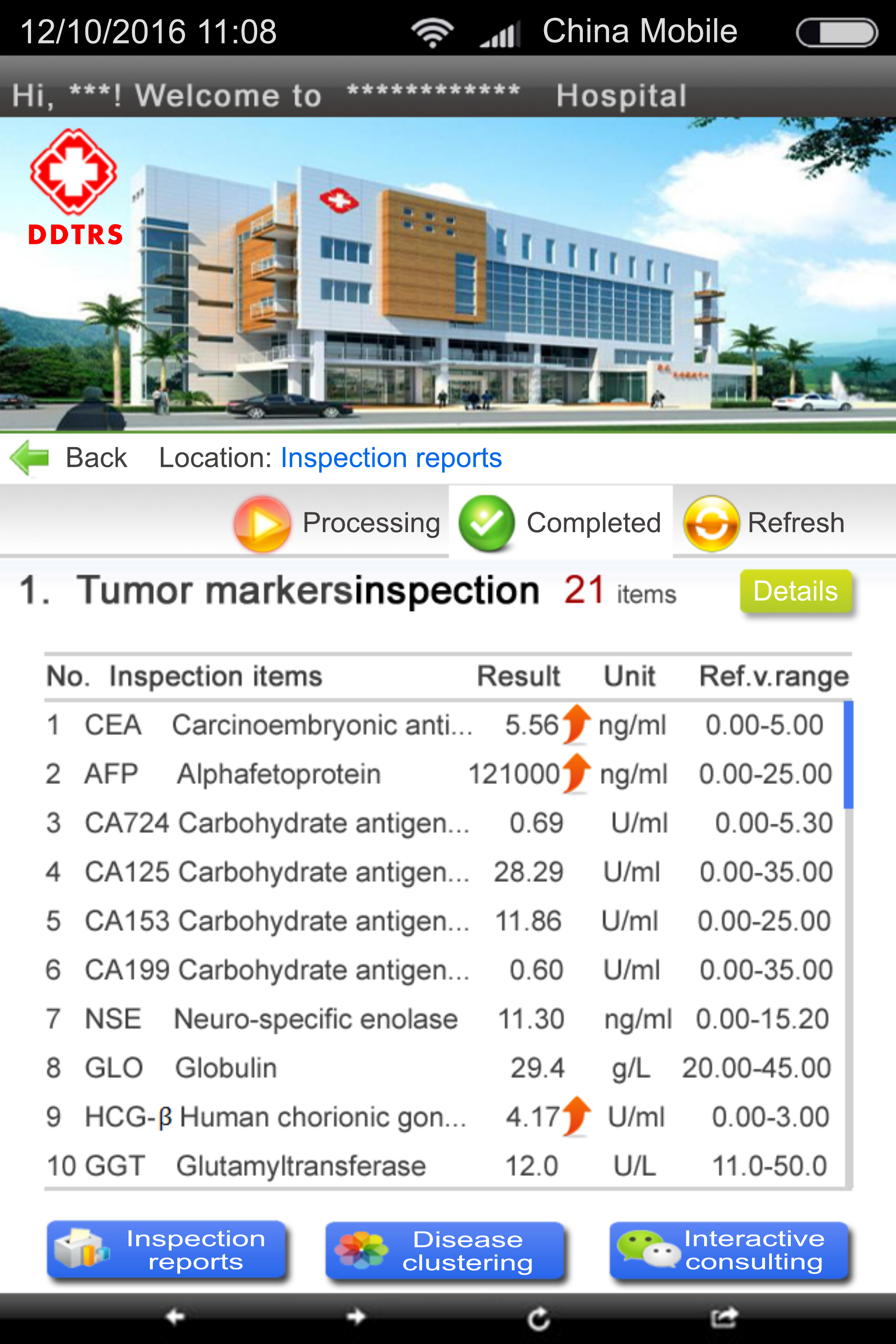}}
 \hspace{0.1in}
 \subfigure[Disease-symptom clustering]{
 \label{fig09:b}
 \includegraphics[width=1.4in]{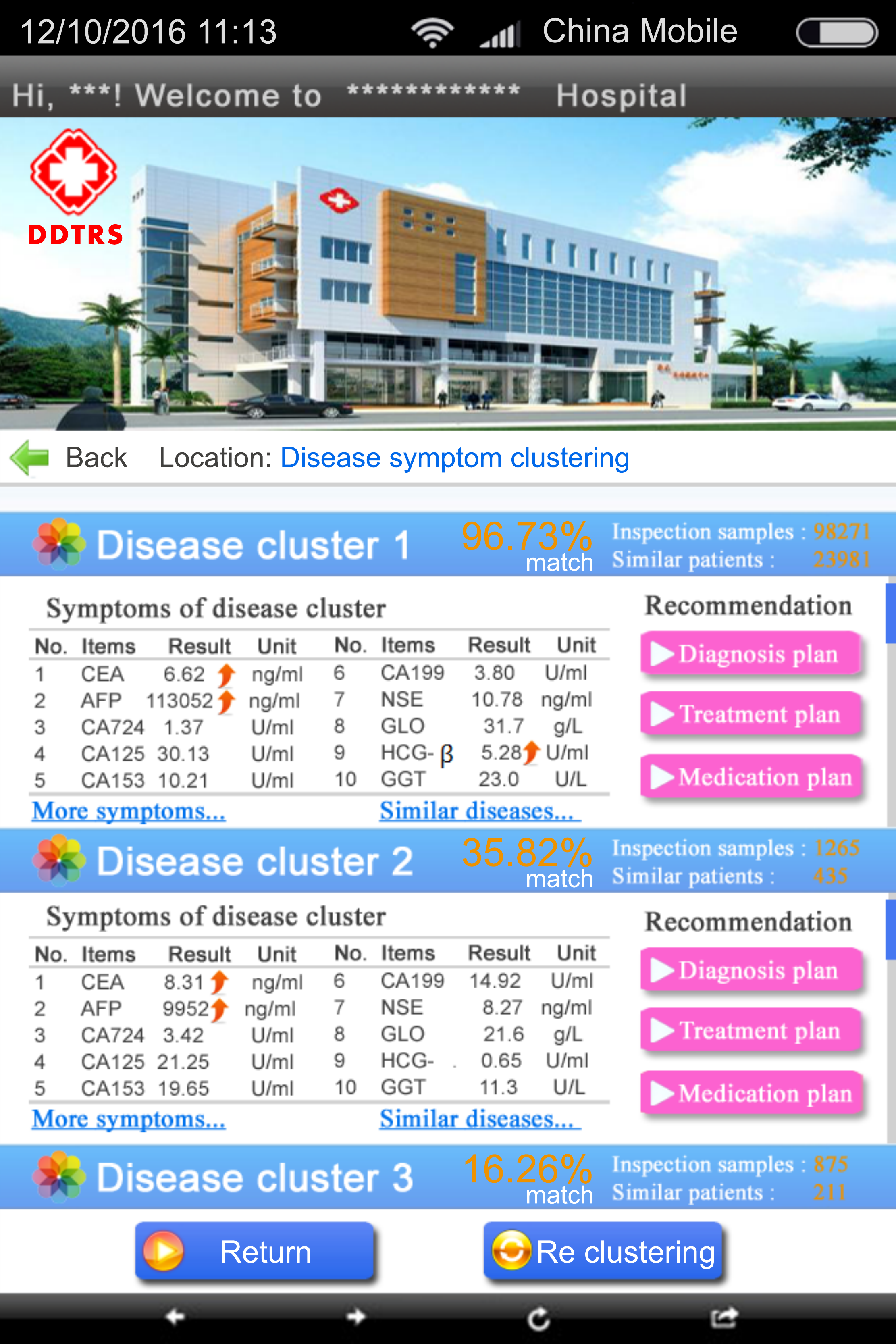}}
  \hspace{0.1in}
 \subfigure[Treatment recommendation]{
 \label{fig09:c}
 \includegraphics[width=1.4in]{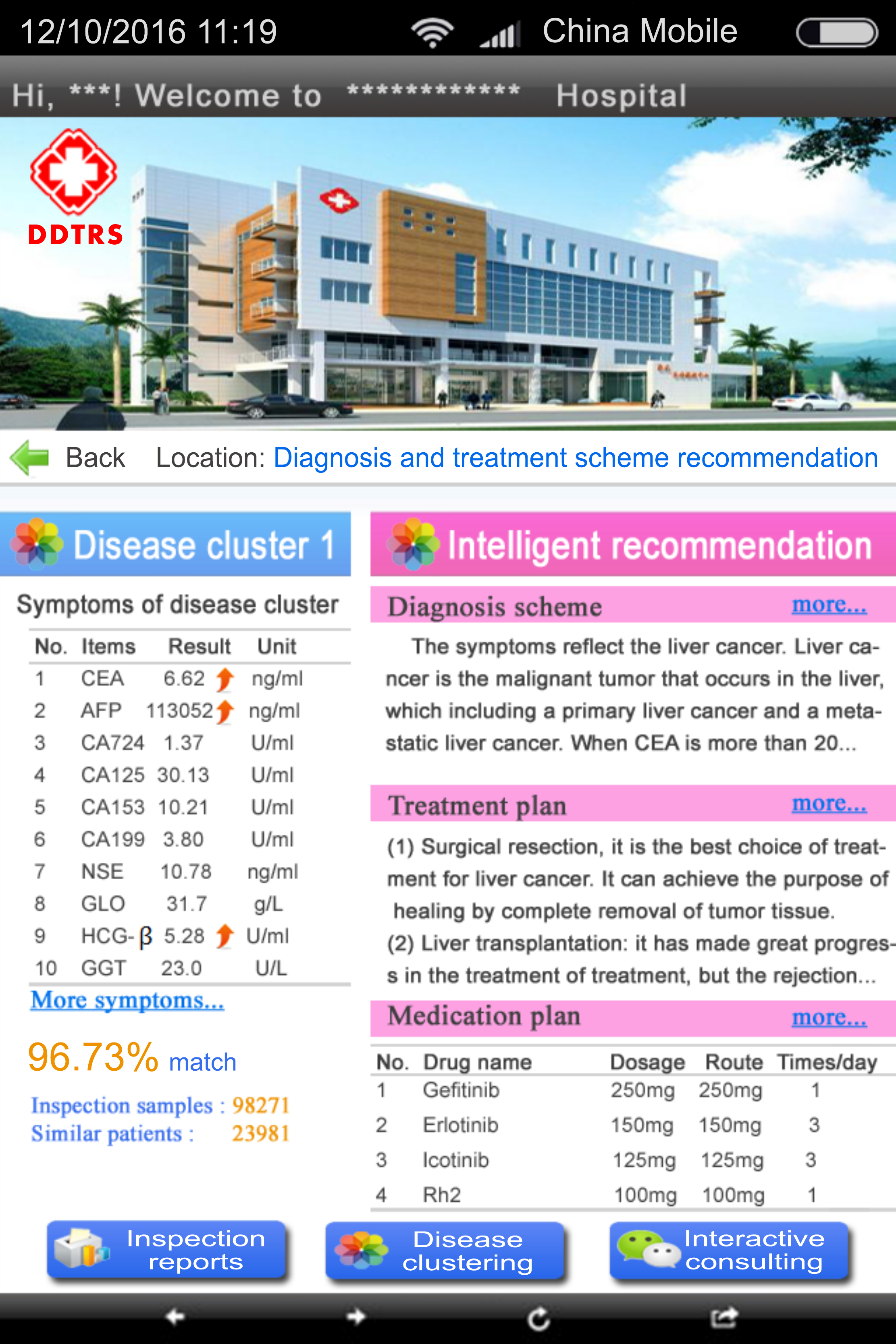}}
  \hspace{0.1in}
 \subfigure[Interactive consulting]{
 \label{fig09:d}
 \includegraphics[width=1.4in]{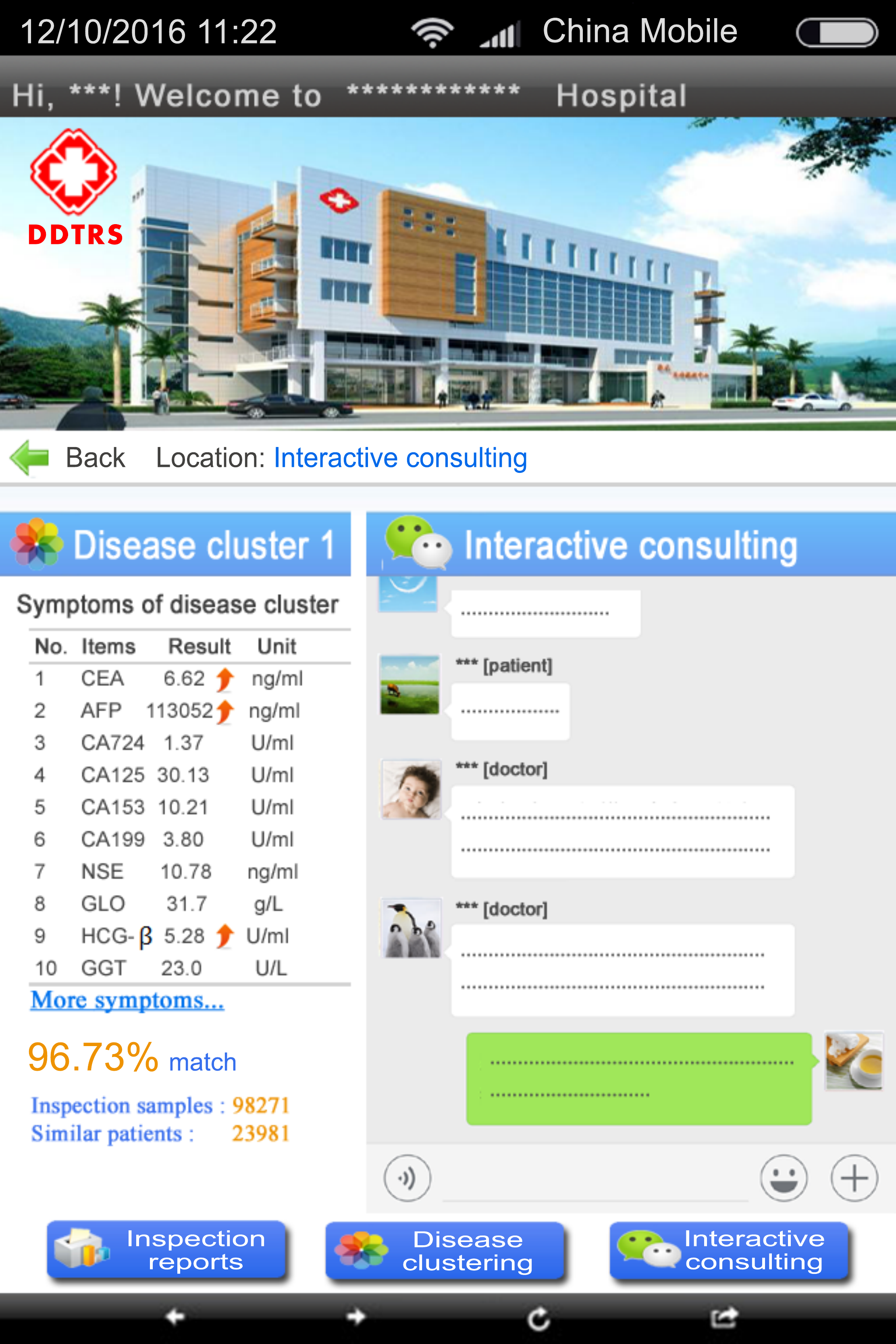}}
 \caption{Mobile recommendation interfaces of DDTRS for patients}
 \label{fig09}
\end{figure*}

A mobile application of DDTRS is provided to patients with interactive recommendation interfaces.
After each visit, when a patient completes the inspection tasks under his medical doctor's preliminary advice, he can obtain the inspection reports through the mobile interfaces.
Examples of interactive mobile recommendation interfaces for patients are shown in Figure \ref{fig09}.

From the recommendation interface, patients can understand and obtain the details of their health conditions via the disease-symptom clustering function.
Moreover, they can further access the corresponding diagnosis and treatment schemes through the interface.
However, there are two cases in which the recommendations might have negative influences on patients.
First, some patients might feel sensitive and anxious when they receive their disease treatment recommendations.
Namely, the recommendations could have bad effects on the patients.
To avoid this problem, before sending the recommendation messages to a patient, an option is provided on the doctor's interface that allows the doctor to determine whether the message should be accessible to the patient.
Second, some patients might have difficulty in understanding the meanings of recommendations and their health conditions because they do not have the medical knowledge that the doctors have.
In this case, they can further consult their doctors or access references from websites.
In consideration of the lack of patient medical knowledge and incomplete feedback, evaluations from patients are not considered in this work.

\section{Parallel Solution of DDTRS}

The performance of disease diagnosis and treatment recommendation is the focus of this section.
The efficiency of the disease-symptom clustering and the latency response of the treatment recommendations are the critical issues of the proposed DDTRS.
After submitting the symptom data of a patient's disease, one expects to receive an appropriate and timely treatment recommendation.
It would be convenient, useful, and preferable if the patients could receive appropriate diagnosis and treatment plans through an interactive mobile application in real time.
Therefore, to reach the goals of high performance and low latency response, we parallelize DDTRS on the Apache Spark cloud computing platform.
The clustering process of disease symptoms and the association analysis process of the disease treatment scheme are parallelized separately.

\subsection{Parallel Clustering Process of Disease Symptoms}

In the parallel solution of the disease-symptom clustering module of DDTRS, large-scale historical inspection datasets are gathered from the cooperating hospitals of this work in a fixed time interval.
DDTRS is deployed on the Spark cloud platform at the National Supercomputing Center in Changsha (NSCC) of China.
Afterwards, these datasets are stored in the Hadoop Distributed File System (HDFS) on the Spark cloud platform.
The parallel clustering process of disease symptoms is implemented on the Spark computing cluster with the RDD programming model.
Apache Spark is a popular parallel data processing platform, that is especially suitable for big data mining and machine learning.
In contrast to Hadoop, parallel programming models of RDD and DAG are supported, which are built on a memory computing framework.
These methods reduce the volume of data transmission operations in the distributed environment without reducing the algorithm's accuracy.

\subsubsection{RDD Dependence for Large-scale Inspection Data}

Before the parallel clustering process, massive volumes of inspection data are loaded into the Tachyon system of the Spark platform with a type of RDD object.
The RDD model is the core programming model of Spark and represents a collection of distributed items.
RDD objects are manipulated in parallel across many computing nodes.
As discussed above, due to different numbers of the inspection items, the dimensions of the datasets for different inspection tasks are different.
The clustering process of each inspection task is executed in parallel separately.
An RDD object $RDD_{task}$ is created for the inspection data subset of each inspection task.
Resulting from the unconstrained clustering process, RDD independencies also occur between different inspection tasks.
Afterwards, each of the RDD objects is allocated to one or multiple adjacent computing nodes of the Spark platform.

In Spark, each RDD object offers two types of operations: (a) $Transformations$ and (b) $Actions$.
$Transformations$ include some operations on RDD objects that return a new RDD object, such as the $map$ and $filter$ functions.
$Actions$ include some operations that compute a result based on an RDD object and return it to the driver program or save it to an external storage system (e.g., HDFS).
At the $Transformation$ stage, $m$ RDD objects are created to store $m$ data subsets of inspection tasks, which are split from the original RDD object $RDD_{Tasks}$ grouping by the name of the inspection task.
In Spark, these objects are computed in a lazy fashion way; namely, they are created in an action at the $Action$ stage.
At the $Action$ stage, $m$ defined RDD objects are created in the Tachyon system.

\begin{figure}[!ht]
\setlength{\abovecaptionskip}{0pt}
\setlength{\belowcaptionskip}{0pt}
\centering
\includegraphics[width=3.5in]{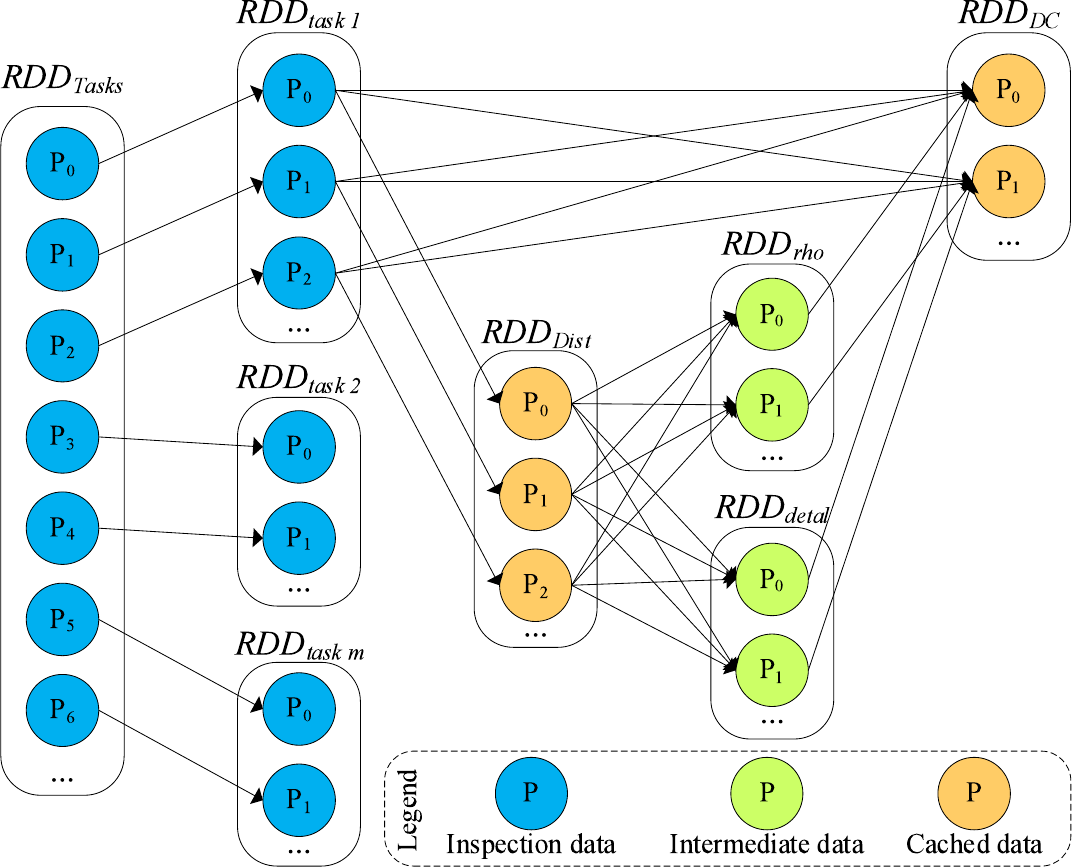}
\caption{RDD dependencies of disease-symptom clustering analysis}
\label{fig10}
\end{figure}

In the subsequent process of disease-symptom clustering analysis, each $RDD_{task}$ is calculated, and multiple new RDD objects are generated from $RDD_{task}$.
Data dependencies occur among the RDD objects generated for the same inspection task in the clustering process.
RDD dependencies of disease-symptom clustering analysis are presented in Figure \ref{fig10}.

In Spark, there are two types of RDD dependencies: (a) narrow dependencies and (b) wide dependencies.
In a narrow dependency, each partition of the parent RDD is used by at most one partition of the child RDD.
In contrast, in a wide dependency, multiple child partitions might depend on one partition of the parent RDD.
As is evident in Figure \ref{fig10}, there are various RDD dependency relationships in the clustering process.
Obviously, a narrow dependency exists between $RDD_{tasks}$ and each inspection data subset $RDD_{task}$.
Afterwards, for each $RDD_{task}$, an $RDD_{Dist}$ object is created to save the distance matrix.
Similarly, a narrow dependency occurs between $RDD_{Dist}$ and $RDD_{task}$.
In the subsequent process of the disease-symptom clustering, objects $RDD_{rho}$ and $RDD_{delta}$ are created to save the values of the local density and the delta distance separately.
Owing to the cross-partition calculation, wide dependencies exist between $RDD_{rho}$ and $RDD_{Dist}$, $RDD_{delta}$ and $RDD_{Dist}$, respectively.
Finally, an $RDD_{DSC}$ object is created to save the disease-symptom clusters, which is calculated based on the $RDD_{rho}$, $RDD_{delta}$, and $RDD_{task}$.
Therefore, wide dependencies occur between $RDD_{DSC}$ and $RDD_{rho}$, $RDD_{delta}$, and $RDD_{task}$.

\subsubsection{Parallel Clustering Process Based on the DAG Model}

According to the RDD dependency, the master computing node of the Spark cluster constructs a task-scheduling DAG for the disease-symptom clustering process.
The parallel clustering process based on the Apache Spark platform is presented in Algorithm \ref{alg2}.
The detailed steps of the process are described as follows.

\begin{algorithm}[!ht]
\caption{Parallel process of the DPCA-based disease-symptom clustering}
\label{alg2}
\begin{algorithmic}[1]
\REQUIRE ~\\
    $Path_{X}$: the path of the inspection datasets stored on HDFS;\\
    $P_{d_{c}}$: the percentage of distance for cutoff distance; \\
    $\varepsilon_{\rho}$: threshold value of the local density for cluster-center decision;\\
    $\varepsilon_{\delta}$: threshold value of the delta distance for cluster-center decision.\\
\ENSURE ~\\
    $RDD_{DSC}$: RDD objects of the disease-symptom clusters.
\STATE $conf$ $\leftarrow$ new SparkConf(``DPCA'', ``SparkMaster'');
\STATE $sc$ $\leftarrow$ new SparkContext($conf$);
\STATE inspection datasets $data$ $\leftarrow$ $sc$.textFile($Path_{X}$);
\STATE $RDD_{Tasks}$ $\leftarrow$ $data$.\textbf{map}
\STATE \quad $fields$ $\leftarrow$ $line$.split(``::'');
\STATE \quad return inspection record ($fields(0), ~fields(1)$);
\STATE \textbf{end map};
\STATE $RDD_{Tasks}$.groupByKey().\textbf{foreach}
\STATE \quad calculate distance matrix $RDD_{Dist}$ from $RDD_{task}$;
\STATE \quad $RDD_{Dist}^{'}$ $\leftarrow$ arrange $RDD_{Dist}$ in ascending order;
\STATE \quad $d_{c}$ $\leftarrow$ distance of the $P_{d_{c}}$ position in $RDD_{Dist}^{'}$;
\STATE \quad ($RDD_{rho}, ~RDD_{delta}$) $\leftarrow$ $RDD_{task}$.\textbf{map}
\STATE \qquad $\rho_{i}$ $\leftarrow$ calculate local density $\sum_{j \in N, j \neq i}{e^{- \left(\frac{d_{ij}}{2d_{c}} \right)^{2}}}$;
\STATE \qquad $\delta_{i}$ $\leftarrow$ calculate delta distance $\min_{j:\rho_{j} < \rho_{i}}{d_{ij}}$;
\STATE \qquad return ($\rho_{i}, ~\delta_{i}$);
\STATE \quad \textbf{end map}.reduce().collect();
\STATE \quad ${DSCc_{s}}$ $\leftarrow$ identify cluster centers($RDD_{rho}, ~RDD_{delta}$).\textbf{map}
\STATE \qquad \textbf{if} ($\rho_{i}$ bigger than $\varepsilon_{\rho}$ and $\delta_{i}$ bigger than $\varepsilon_{\delta}$) \textbf{then}
\STATE \qquad \quad ${DSC_{c}}$ $\leftarrow$ identify cluster centers ($\rho_{i}, ~\delta_{i}$);
\STATE \qquad \quad return $DSC_{c}$;
\STATE \qquad \textbf{end if}
\STATE \quad \textbf{end map}
\STATE \quad $RDD_{DSC}$ $\leftarrow$ $RDD_{task}$.\textbf{flatMap}
\STATE \qquad append $x_{i}$ to the nearest cluster $DSC_{n}$;
\STATE \quad \textbf{end flatMap}.reduceByKey();
\STATE \textbf{end foreach}.collect();
\RETURN $RDD_{DSC}$.
\end{algorithmic}
\end{algorithm}

(1) The execution environment of the Spark platform is configured, such as the name of the program and the address of the Spark platform.
Then, massive volumes of historical inspection datasets are loaded from HDFS to the Spark Tachyon memory system as an RDD object (termed $RDD_{Tasks}$).
In the $groupByKey$ function, records of $RDD_{Tasks}$ are divided into a series of RDD objects (each one is termed $RDD_{task}$) by inspection task name.

(2) In the $foreach$ parallel function, the disease-symptom clustering process of each $RDD_{task}$ is executed simultaneously.
An object $RDD_{Dist}$ is created to save the value of a distance matrix from the current $RDD_{task}$.
The local density $\rho_{i}$ and delta distance $\delta_{i}$ of each record $x_{i}$ are measured in parallel in the $map$ function.
Then, the corresponding objects $RDD_{rho}$ and $RDD_{delta}$ of $RDD_{task}$ are obtained in the $reduce$ and $collect$ functions.

(3) Records with high values of both $RDD_{rho}$ and $RDD_{delta}$ are identified as the centers of disease-symptom clusters (termed $DSCc_{s}$).
Afterwards, each of the remaining inspection records in $RDD_{task}$ is assigned to the nearest cluster in the $flatMap$ parallel function.
Owing to the independence of $RDD_{task}$ during the remaining-record assignment step, this step is executed in parallel.
Moreover, there is no data transmission consumption among the computing nodes that the dataset located.
Finally, an RDD object $RDD_{DSC}$ is obtained for the detected disease-symptom clusters.

\subsection{Parallel Association Analysis of D-T Rules}

To increase the speed of disease treatment-scheme association analysis and achieve a low latency response of the treatment recommendation, we parallelize the association analysis process on the Apache Spark cloud computing platform.

\subsubsection{RDD Dependence for Large-scale Treatment Data}

\begin{figure}[!ht]
\setlength{\abovecaptionskip}{0pt}
\setlength{\belowcaptionskip}{0pt}
\centering
\includegraphics[width=3.3in]{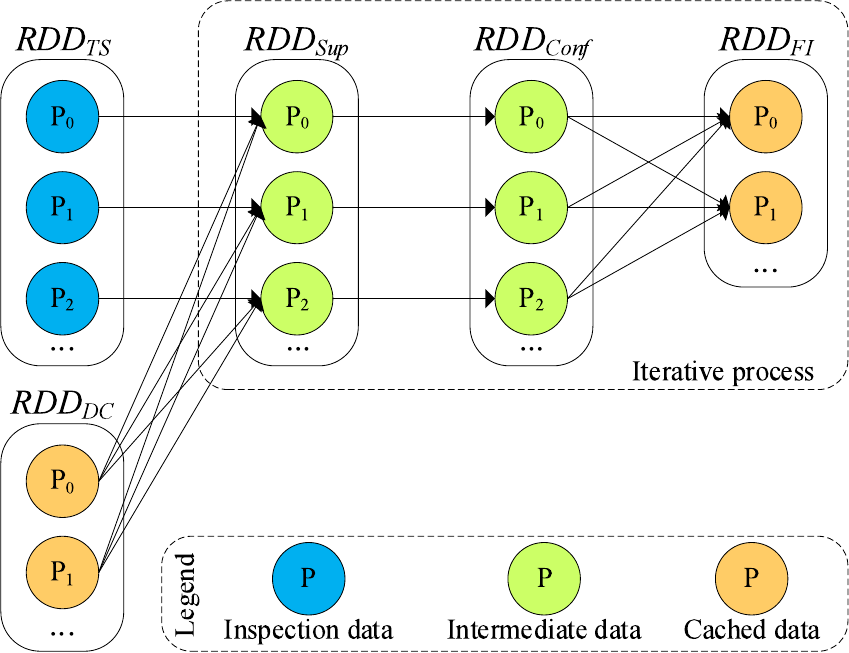}
\caption{RDD dependencies of the treatment-scheme association analysis}
\label{fig11}
\end{figure}

Analogous to the inspection data, large-scale historical treatment-scheme datasets are gathered from the cooperating hospitals and stored on HDFS in a fixed time interval.
Before the association analysis process, the treatment-scheme datasets are loaded into the Spark Tachyon memory system with a type of RDD object (termed $RDD_{TS}$).
At the same time, the results of disease-symptom clustering obtained from Algorithm \ref{alg2} are loaded into the Tachyon system as an RDD object $RDD_{DSC}$.
In the subsequent process of disease treatment association analysis, records of $RDD_{TS}$ and $RDD_{DSC}$ objects are calculated.
In addition, multiple new RDD objects are generated according to the data dependency between $RDD_{TS}$ and $RDD_{DSC}$.
RDD dependencies of the treatment-scheme association analysis are presented in Figure \ref{fig11}.

In Figure \ref{fig11}, there are various RDD dependency relationships in the association analysis process.
The RDD object $RDD_{sup}$ of the support value is created based on treatment schemes $RDD_{TS}$ and disease-symptom clusters $RDD_{DC}$.
Because each partition of $RDD_{TS}$ is used by at most one partition of $RDD_{sup}$, a narrow dependency exists between $RDD_{TS}$ and $RDD_{sup}$.
In contrast, since multiple partitions of the child object $RDD_{sup}$ are dependent on one partition of $RDD_{DC}$, a wide dependency exists between $RDD_{DC}$ and $RDD_{sup}$.
In the subsequent process of the association analysis, to obtain the confidence value, an object $RDD_{conf}$ is created from $RDD_{sup}$, with a narrow dependency between them.
Finally, an object $RDD_{FI}$ is created to save the frequent items based on the wide dependency of $RDD_{conf}$.

\subsubsection{Parallel Association Analysis Process of Disease-Treatment Rules}

Similar to the parallel clustering process, the execution environment is configured for the parallel association analysis process.
According to the RDD dependencies in Figure \ref{fig11}, for each disease-symptom cluster $DSC_{i}$, the association analysis process of the corresponding treatment schemes is executed in parallel in the $foreach$ function.
The association rules between $DSC_{i}$ and the related treatment schemes $DSC_{i}.TSs$ are extracted.
In the $flatMap$ parallel function, the support value $RDD_{Sup}$ of each association rule is measured by the Apriori algorithm.
If $RDD_{Sup}$ of an association rule is larger than the minimum support $MinSup$, the confidence value $RDD_{Conf}$ of the rule is derived.
If $RDD_{Conf}$ is larger than the minimum confidence $MinConf$, the current association rule is marked as a strong rule.
In the $flatMap$ and $foreach$ functions, strong rules of disease-symptom clusters $RDD_{DSC}$ are appended to the frequent itemset $RDD_{FI(DSC, ~TS)}$ simultaneously.
Hence, strong association rules of disease treatment schemes are produced.
The parallel process of the Apriori-based disease treatment-scheme association analysis is presented in Algorithm \ref{alg4}.

\begin{algorithm}[!ht]
\caption{Parallel process of the Apriori-based disease treatment-scheme association analysis}
\label{alg4}
\begin{algorithmic}[1]
\REQUIRE ~\\
    $DSCs$: the disease-symptom clusters obtained from Algorithm \ref{alg2};\\
    $Path_{TS}$: the path of the treatment-scheme records stored on HDFS;\\
    $MinSup$: the preset value of the minimum support;\\
    $MinConf$: the preset value of the minimum confidence.\\
\ENSURE ~\\
    $RDD_{FI(DSC, ~TS)}$: the frequent itemsets of the association rules.
\STATE $conf$  $\leftarrow$ new SparkConf(``Apriori'', ``SparkMaster'');
\STATE $sc$  $\leftarrow$ new SparkContext($conf$);
\STATE $RDD_{TS}$  $\leftarrow$  $sc$.textFile($Path_{TS}$);
\STATE create frequent itemset $FI(DSC, ~TS)$;
\STATE sc.parallelize($DSCs$).\textbf{foreach}
   \STATE \quad get disease-symptom cluster $DSC_{i}$;
   \STATE \quad $RDD_{FI(DSC, ~TS)}$ $\leftarrow$ $DSC_{i}.TSs$.\textbf{flatMap}
        \STATE \qquad get treatment scheme $TS_{j}$;
        \STATE \qquad $R_{ix}$ $\leftarrow$ get treatment association $E(DSC_{i}, ~TS_{j})$;
        \STATE \qquad $R_{iy}$ $\leftarrow$ get treatment association $E(DSC_{i}, ~TS_{j+1})$;
        \STATE \qquad calculate the support value $RDD_{Sup}(R_{ix} \Rightarrow R_{iy})$;
        \STATE \qquad \textbf{if} {$RDD_{Sup}(R_{ix} \Rightarrow R_{iy}) \geq MinSup$} \textbf{then}
            \STATE \qquad \quad calculate confidence $RDD_{Conf}(R_{ix} \Rightarrow R_{iy})$;
            \STATE \qquad \quad \textbf{if} {$RDD_{Conf}(R_{ix} \Rightarrow R_{iy}) \geq MinConf$} \textbf{then}
                \STATE \qquad \qquad $RDD_{FI(DSC, ~TS)}$ $\leftarrow$ append frequent itemset ${\{}R_{ix}, ~R_{iy}{\}}$;
             \STATE \qquad \quad \textbf{end if}
        \STATE \qquad \textbf{end if}
   \STATE \quad \textbf{end flatMap}.reduce();
\STATE \textbf{end foreach}.collect().groupBy($DSC$);
\RETURN $RDD_{FI(DSC, ~TS)}$.
\end{algorithmic}
\end{algorithm}

The computational complexity of Algorithm \ref{alg4} is $O(\frac{|c| \overline{k}}{p})$, where $|c|$ is the number of disease-symptom clusters, $\overline{k}$ is the average number of treatment records for each disease-symptom cluster, and $p$ is the number of computing nodes.
Benefiting from the distributed data allocation and the parallel computing mechanism, the time complexity of the Apriori-based disease treatment-scheme association analysis is obviously reduced.

\section{Experiments and Applications}
We evaluate our proposed model in terms of clustering accuracy, recommendation quality, and performance of the proposed DDTRS.
Section \ref{section5.1_ExperimentSetup} presents the experimental settings.
Clustering accuracy and recommendation quality evaluations of DDTRS are presented in Section \ref{section5.2_AccuracyEvaluation}.
Section \ref{section5.3_CaseAnalysis} analyzes two cases of the disease treatment recommendations.
A performance assessment of DDTRS is provided in Section \ref{section5.4_PerformanceEvaluation}.

\subsection{Experimental Setup}
\label{section5.1_ExperimentSetup}

DDTRS is developed in a client/server software model, including a data-collection terminal (client) and a data-analysis terminal (server).
The data-collection terminal is deployed in our cooperating hospitals.
Using the data-collection terminal, massive volumes of historical and current medical datasets are gathered, and the results of clustering analysis and treatment recommendations are fed back to the medical doctors in these hospitals.
The data-analysis terminal is deployed at the National Supercomputing Center in Changsha (NSCC) for the disease-symptom clustering, association analysis, and treatment recommendation.

The experimental setup of DDTRS is installed on the Apache Spark cloud platform at the NSCC, which is comprised of 30 computing nodes.
Each node has eight Intel Xeon Nehalem EX CPU (8 cores, 2.27GHz) and 32GB memory.
The nodes are connected by a high-speed Gigabit network.
Each node is configured with Ubuntu 15.10 and a cloud computing environment using Apache Spark 1.6.0.

\subsection{Accuracy Evaluation}
\label{section5.2_AccuracyEvaluation}

\subsubsection{Accuracy of Disease-Symptom Clustering}

The accuracy of the disease-symptom clustering is a crucial issue in medical sciences.
Inaccurate results of the disease-symptom clustering might lead to incorrect diagnoses and inappropriate treatment recommendations, endangering patients' health.
Considering that predicting disease for a given inspection result is a standard classification issue, in this experiment, two typical classification algorithms (namely, C4.5 and Random Forest (RF)) and a typical clustering algorithm (namely, K-Means) are introduced as the comparison algorithms.
The accuracy of the proposed DPCA-based disease-symptom clustering algorithm is assessed by comparing the results with those of the C4.5, RF, and K-Means algorithms.

\begin{figure}[!ht]
\setlength{\abovecaptionskip}{0pt}
\setlength{\belowcaptionskip}{0pt}
\centering
\includegraphics[width=3.5in]{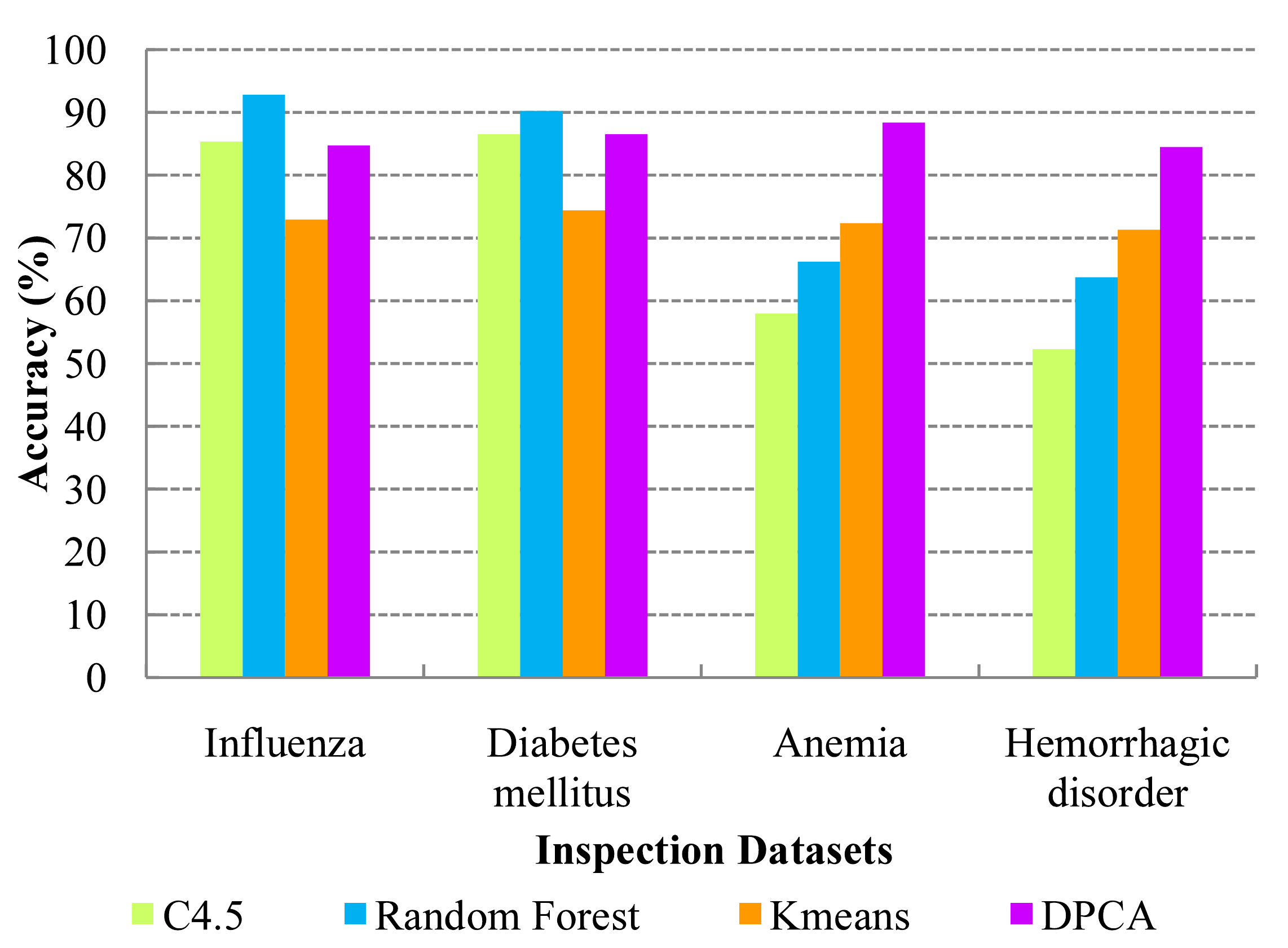}
\caption{Accuracy evaluation of different algorithms for disease-symptom clustering}
\label{chart01}
\end{figure}

Although the clustering algorithm is a kind of unsupervised learning algorithm, for which it is not necessary to label the samples in advance, we pre-defined the class labels for all the samples in this experiment for comparison with the classification algorithms.
Cluster Accuracy (CA) is introduced to evaluate the clustering algorithms and the classification algorithms \cite{cc013}.
CA measures the ratio of the number of correctly classified / clustered instances to that of pre-defined class labels.
Let $X$ be the inspection dataset in this experiment, let $C$ be the set of classes / clusters detected by the corresponding algorithm, and let $L$ be the set of pre-defined class labels.
CA is defined in Eq. (\ref{eq15}):

\begin{equation}
\label{eq15}
CA = \sum_{i=0}^{K-1}{\frac{\max{(C_{i}|L_{i})}}{|X|}},
\end{equation}
where $C_{i}$ is the set of data points in the $i$-th class/cluster, $L_{i}$ is the set of pre-defined class labels of the data points in $C_{i}$, and $K$ is the size of $C$.
$max(C_{i}|L_{i})$ is the number of data points that have the majority label in $C_{i}$.
The greater value of $CA$, the higher the accuracy of the classification / clustering algorithm and the greater the purity that each cluster achieves.
The comparison results of disease-symptom clustering based on different algorithms are illustrated in Figure \ref{chart01}.

As is evident from Figure \ref{chart01}, for diseases that have few treatment stages or pathogeneses, such as influenza and Diabetes Mellitus (DM), the traditional classification algorithms obtain higher accuracy than the clustering algorithms.
For example, in the case of influenza, the accuracy of C4.5 is 85.32\% and that of RF is 92.82\%, and those of the K-Means and DPCA algorithms are 70.92\% and 84.73\%, respectively.
In the case of DM, the accuracies of C4.5 and RF are 86.48\% and 90.23\%, while those of K-Means and DPCA are 74.38\% and 86.53\%, respectively.
This is because there are fewer treatment stages of diseases and fewer changes in the disease symptoms.
In contrast, for diseases with the characteristics of multiple treatment stages, various symptoms, or multi-pathogenesis, the clustering algorithms achieve higher degrees of accuracy than the traditional classification algorithms.
For example, in the case of anemia, the accuracy of DPCA is higher than those of the comparison algorithms, peaking at 88.35\%, while those of the K-Means, RF, and C4.5 algorithms are 72.36\%, 66.21\%, and 57.92\%, respectively.
The experimental results demonstrate that our DPCA-based disease-symptom clustering algorithm achieves stability and high accuracy.

\subsubsection{Robustness of the DPCA Algorithm}

To intuitively describe the DPCA clustering process of disease symptoms, we take a dataset of the tumor-marker inspection task as an example.
The disease-symptom clustering analysis of this dataset is expected to identify whether the patients corresponding to these data have a liver cancer.
Liver cancer is a malignant tumor that occurs in the liver, which includes a primary liver cancer and metastatic liver cancer.
In the tumor-marker inspection task, there are more than 10  inspection items, such as alpha-fetoprotein (AFP), carcinoembryonic antigen
(CEA), carbohydrate antigen (CA724), CA125, CA153, CA199, neuro-specific enolase (NSE), globulin (GLO), human chorionic gonadotophin-$\beta$ (HCG-$\beta$), and gamma glutamyl transpeptidase (GGT).
Among them, AFP and CEA are the most crucial inspection items for identifying liver cancer.
AFP measurement is one of the most specific methods for the diagnosis of hepatocellular carcinoma.
CEA is a tumor index of digestive tract cancer, which particularly deteriorates and transforms into liver cancer.
When AFP is more than 25 \emph{ng/ml} and CEA is more than 5 \emph{ng/ml}, a tumor is often suggested, which is common in liver cancer, colorectal cancer, and breast cancer.

To facilitate expression and understanding, a dataset of the tumor-marker inspection task containing 40 samples is collected, and two inspection items (namely, AFP and CEA) of the inspection task are selected to construct a two-dimensional data view.
Afterwards, we analyze the calculation processes, such as distance calculation and density peak calculation.
The dataset of the tumor-marker inspection task is shown in Table \ref{table04}.

\begin{table}[!ht]
\caption{Dataset of the tumor-marker inspection task (partial)}
\centering
\label{table04}
\begin{tabular}{cccccccccccc}
\hline
No.	& AFP	& CEA       & No.	& AFP	& CEA       & No.	& AFP	& CEA       & No.	& AFP	& CEA \\
\hline
1	& 1.06 	& 10.70 	& 11	& 6.24 	& 19.42 	& 21	& 4.75 	& 26.85 	& 31	& 6.34 	& 27.11 \\
2	& 2.32 	& 13.50     & 12	& 7.38 	& 10.35 	& 22	& 4.91 	& 27.98 	& 32	& 6.34 	& 25.32 \\
3	& 4.55 	& 20.45 	& 13	& 8.00 	& 8.10   	& 23	& 4.82 	& 25.48 	& 33	& 6.03 	& 23.37 \\
4	& 4.68 	& 18.23 	& 14	& 3.57 	& 27.73 	& 24	& 4.24 	& 23.32 	& 34	& 6.65 	& 27.37 \\
5	& 5.01 	& 19.94 	& 15	& 3.78 	& 25.06 	& 25	& 4.70 	& 22.65 	& 35	& 6.60 	& 25.83 \\
6	& 5.11 	& 18.14 	& 16	& 4.08 	& 25.93 	& 26	& 5.30 	& 27.62 	& 36	& 6.60 	& 24.29 \\
7	& 5.32 	& 20.96 	& 17	& 4.50 	& 27.26 	& 27	& 5.11 	& 26.09 	& 37	& 6.65 	& 21.73 \\
8	& 5.37 	& 19.42 	& 18	& 4.54 	& 26.03 	& 28	& 5.21 	& 24.04 	& 38	& 6.96 	& 25.06 \\
9	& 5.62 	& 17.89 	& 19	& 4.70 	& 22.65 	& 29	& 5.73 	& 28.65 	& 39	& 7.11 	& 26.60 \\
10	& 5.73 	& 20.19 	& 20	& 4.65 	& 28.14 	& 30	& 5.88 	& 26.09 	& 40	& 7.47 	& 22.75 \\
\hline
\end{tabular}
\end{table}

\begin{figure}[!ht]
\setlength{\abovecaptionskip}{0pt}
\setlength{\belowcaptionskip}{0pt}
\centering
 \subfigure[Eq. (\ref{eq02}) ($d_{c}=1.0\%$)]{
 \label{chart02:a}
 \includegraphics[width=1.5in]{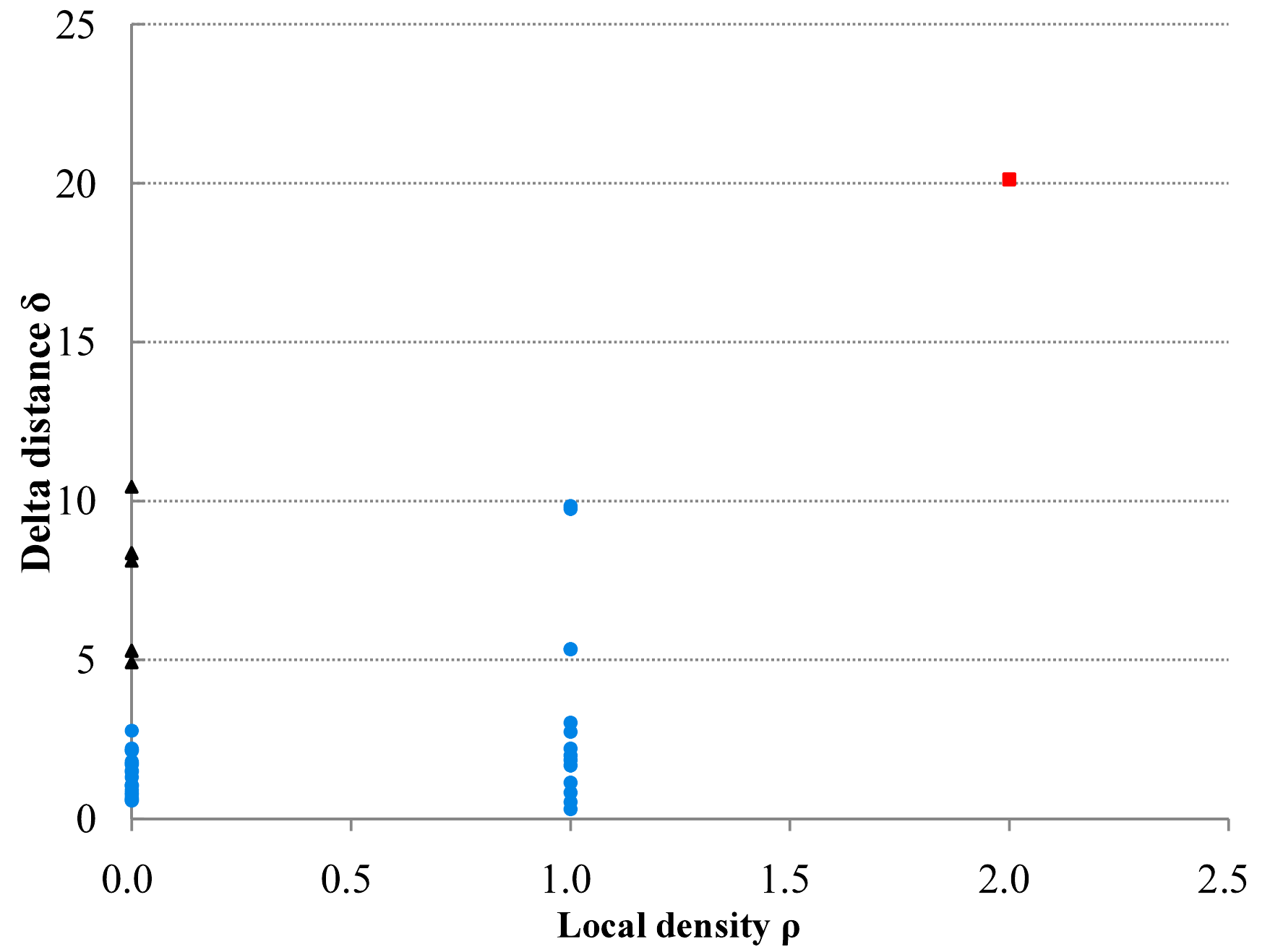}}
 \subfigure[Eq. (\ref{eq02}) ($d_{c}=2.0\%$)]{
 \label{chart02:b}
 \includegraphics[width=1.5in]{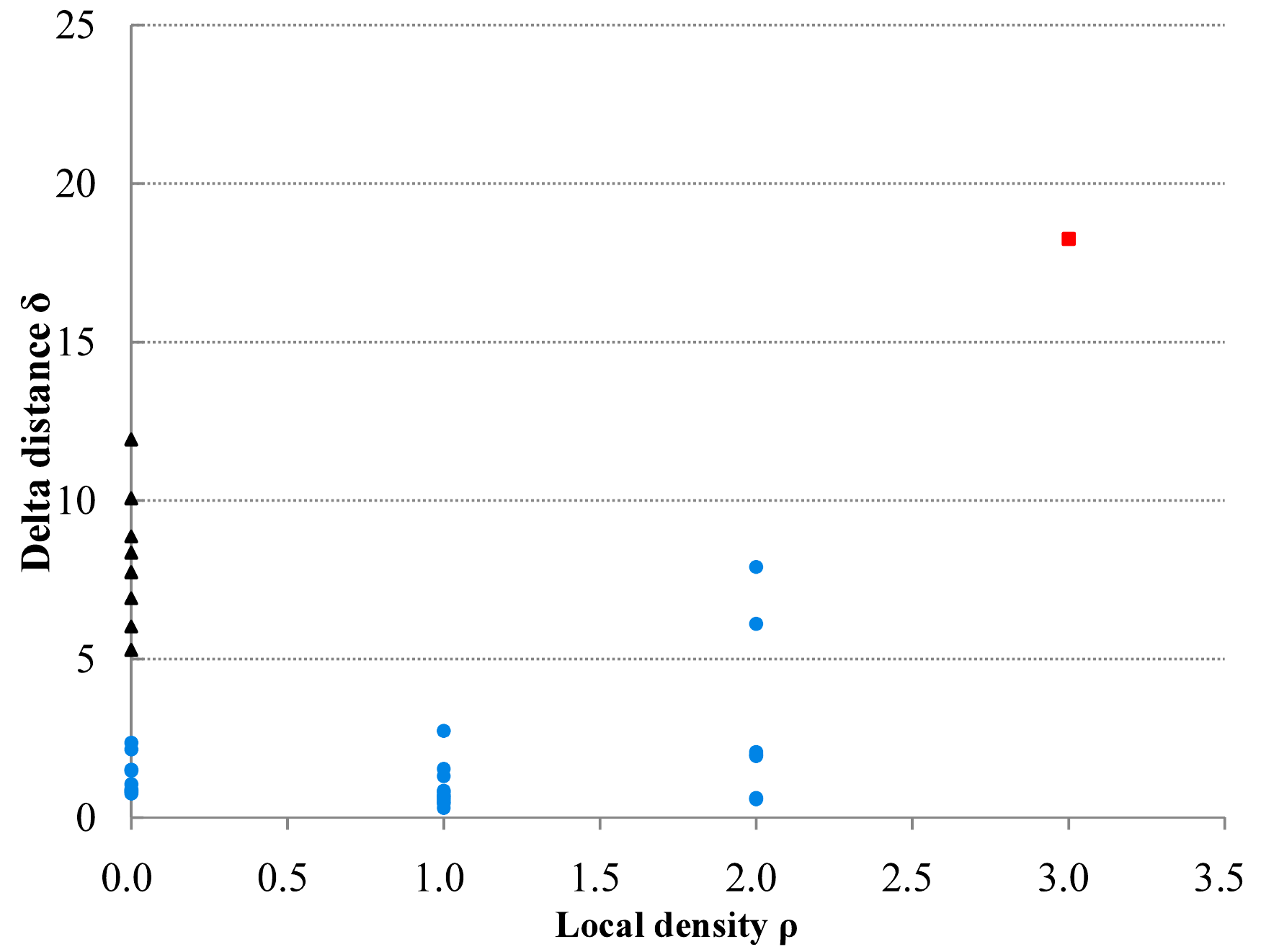}}
 \subfigure[Eq. (\ref{eq02}) ($d_{c}=3.0\%$)]{
 \label{chart02:c}
 \includegraphics[width=1.5in]{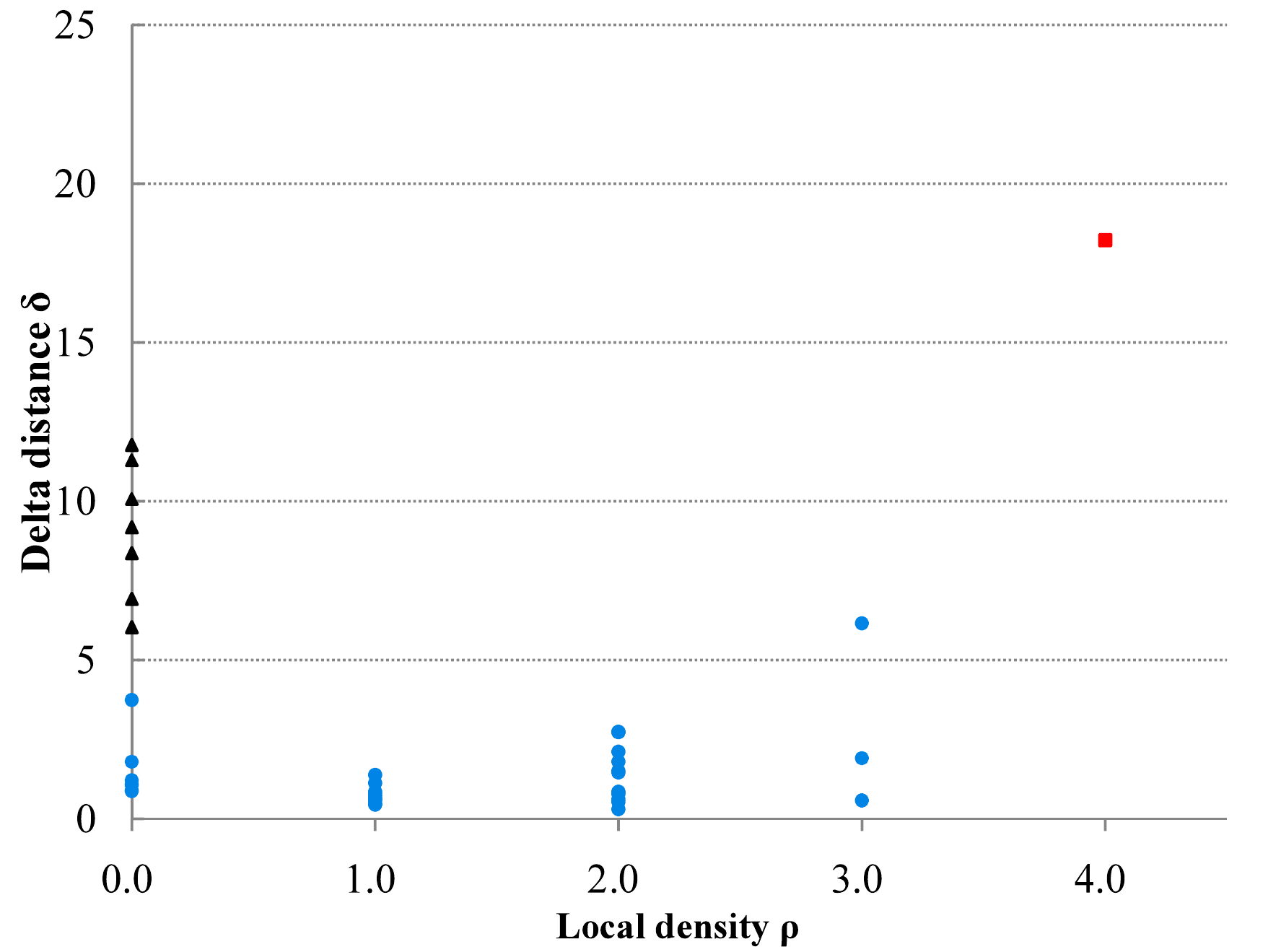}}
 \subfigure[Eq. (\ref{eq02}) ($d_{c}=4.0\%$)]{
 \label{chart02:d}
 \includegraphics[width=1.5in]{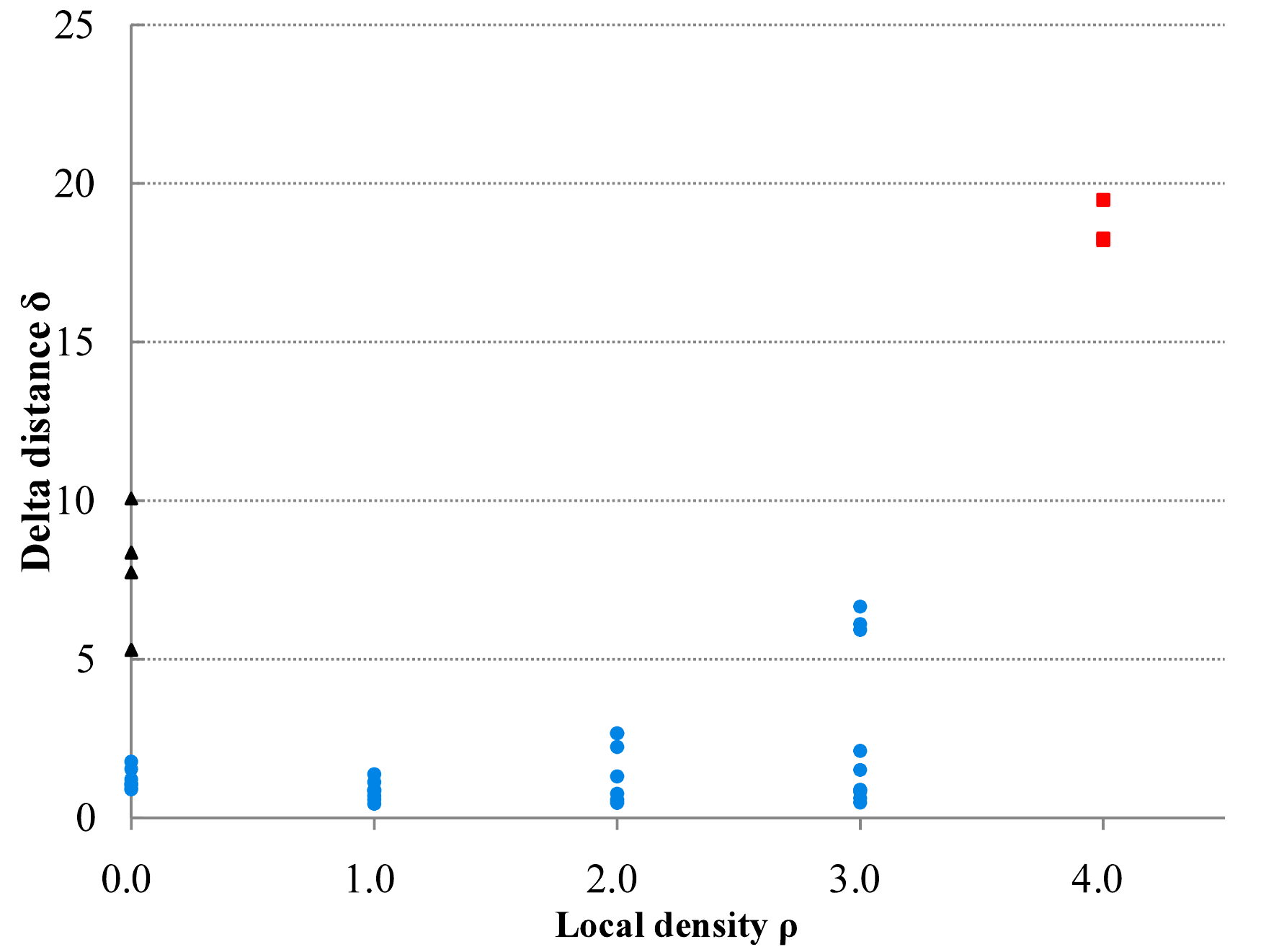}}\\
  \subfigure[Eq. (\ref{eq04}) ($d_{c}=1.0\%$)]{
 \label{chart02:e}
 \includegraphics[width=1.5in]{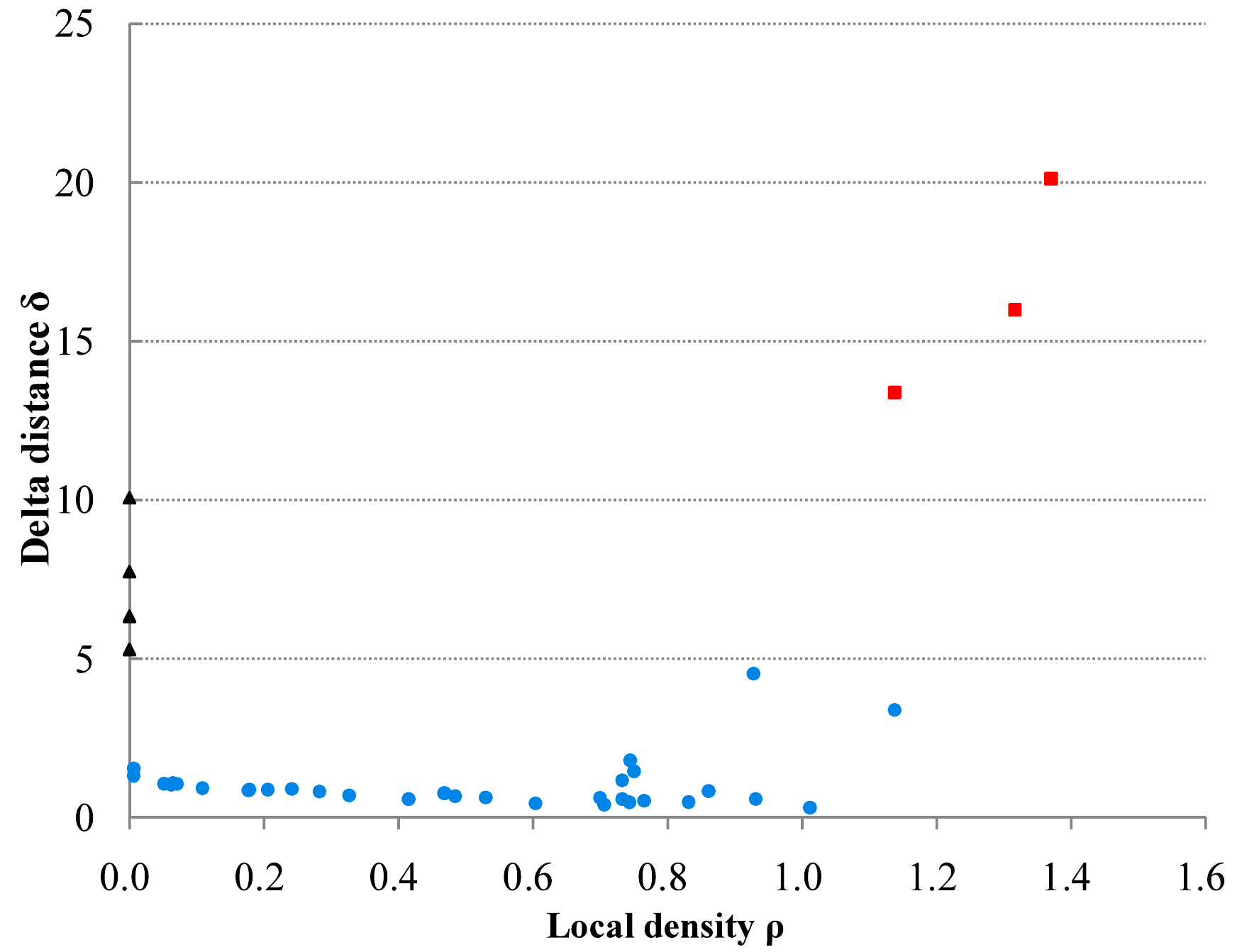}}
 \subfigure[Eq. (\ref{eq04}) ($d_{c}=2.0\%$)]{
 \label{chart02:f}
 \includegraphics[width=1.5in]{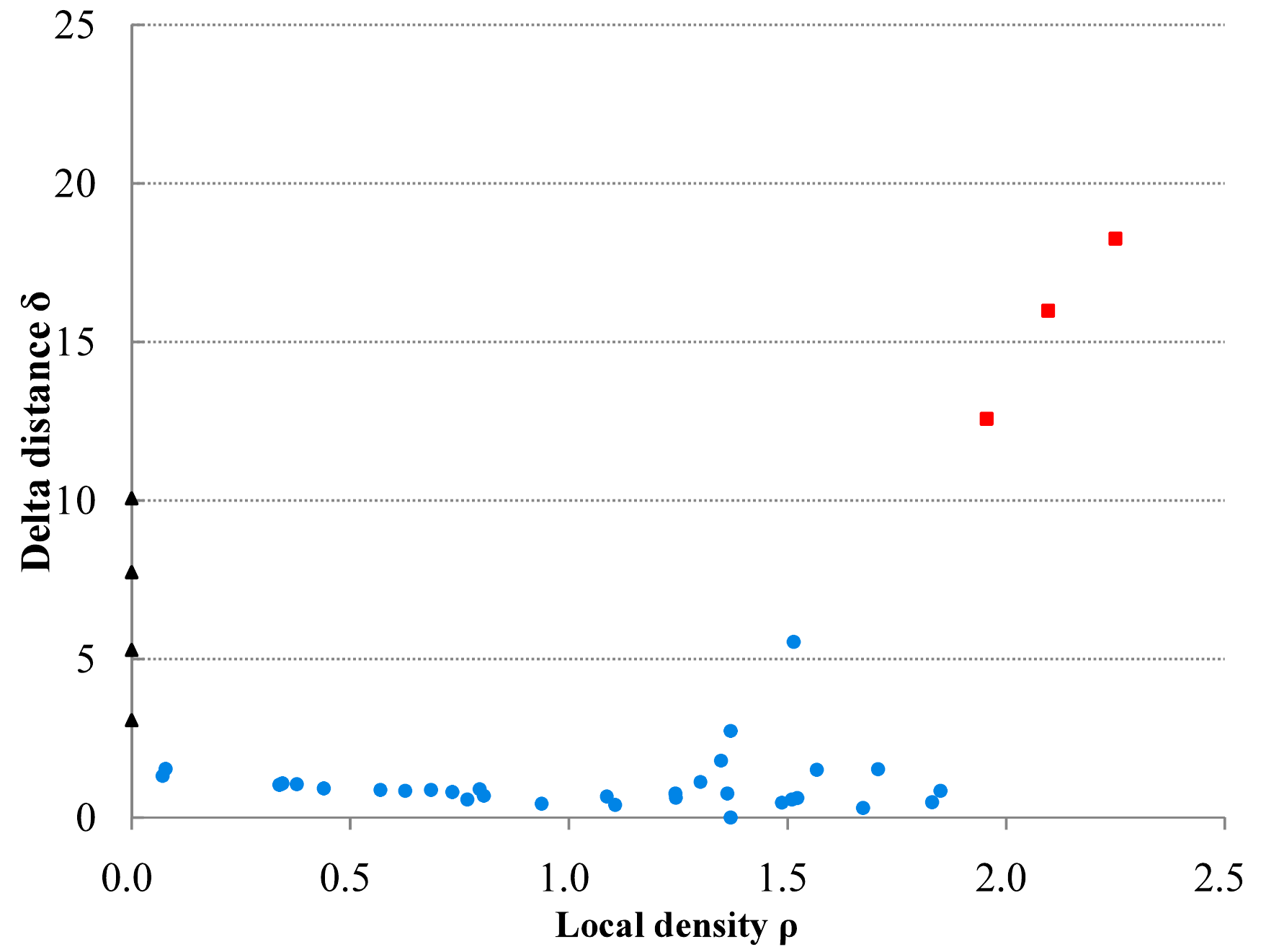}}
 \subfigure[Eq. (\ref{eq04}) ($d_{c}=3.0\%$)]{
 \label{chart02:g}
 \includegraphics[width=1.5in]{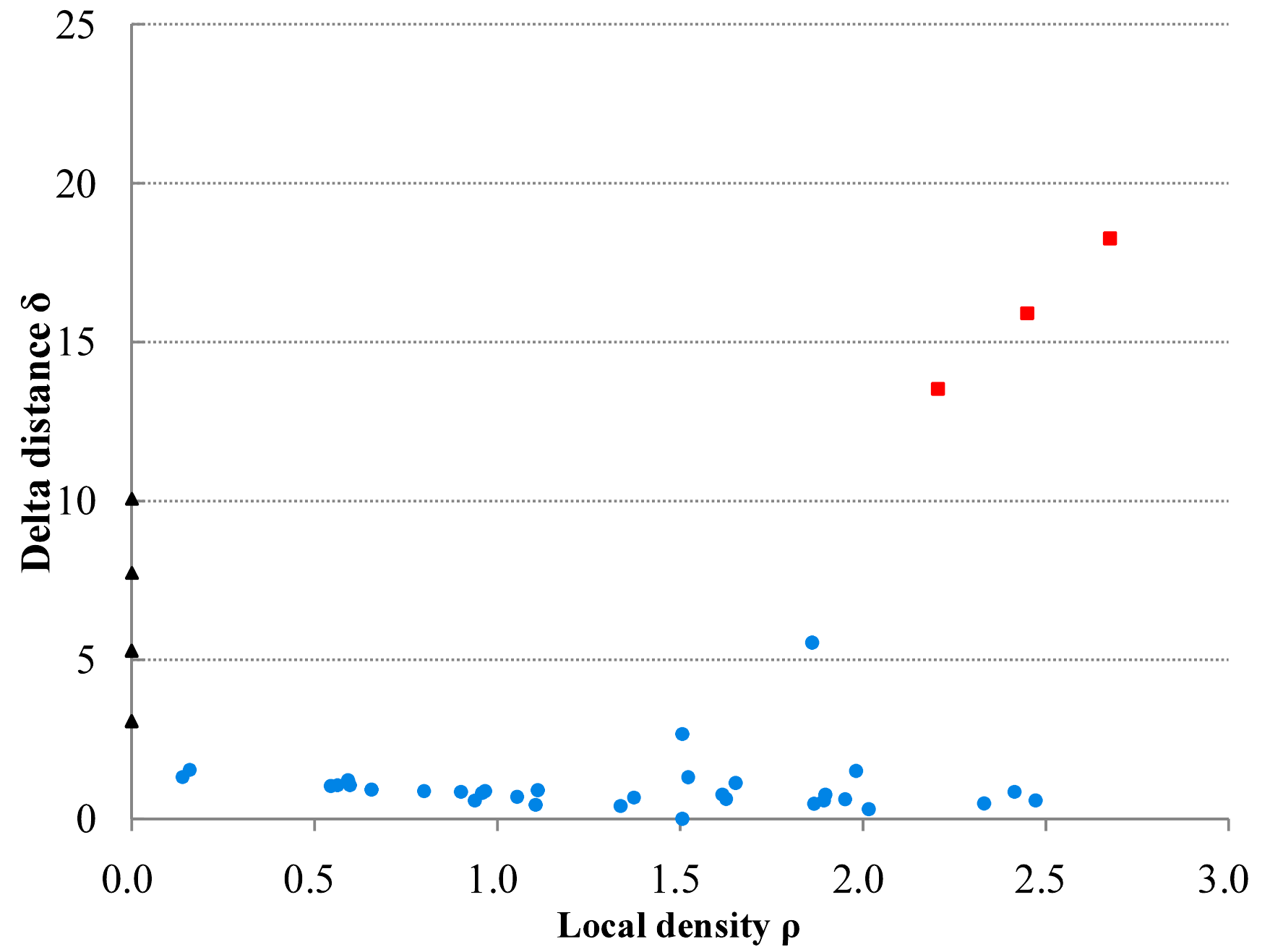}}
 \subfigure[Eq. (\ref{eq04}) ($d_{c}=4.0\%$)]{
 \label{chart02:h}
 \includegraphics[width=1.5in]{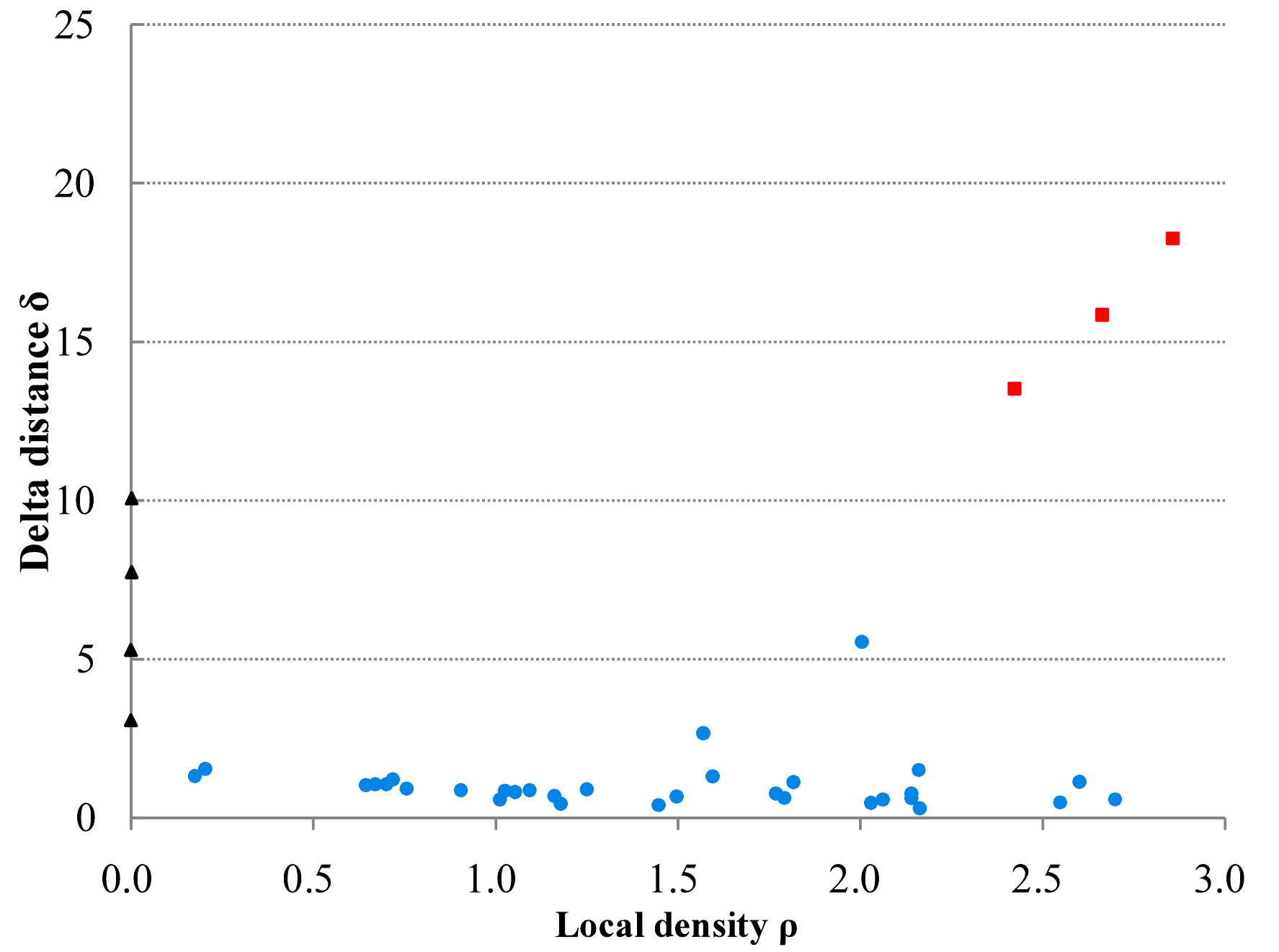}}
 \caption{Comparison of decision graphs obtained by Eq. (\ref{eq02}) and Eq. (\ref{eq04})}
 \label{chart02}
\end{figure}

To compare the effects of different calculation methods of local density on the robustness of the algorithm, we set different values of the cutoff distance in the experiment.
For the 40 data points of the dataset in Table \ref{table04}, there are 780 distance values between data points, which are calculated by Eq. (\ref{eq01}).
These distances are arranged in ascending order, and the cutoff distance $d_{c}$ is set to the distance values of the 1.0\%, 2.0\%, 3.0\%, and 4.0\% positions of the ordered distances.
Namely, the cutoff distance $d_{c}$ is set as 0.57,  0.76, 0.85, and 0.88, respectively.
The influence of the cutoff distance on the local density is evaluated by analyzing the variation of the data density under different cutoff distances.
We calculate the local density $\rho$ for the dataset by Eq. (\ref{eq02}) and Eq. (\ref{eq04}) depending on different cutoff distances.
Moreover, corresponding delta distances $\delta$ are subsequently calculated based on the related local density $\rho$.
The results of comparing the effects of different calculation methods of local density are presented in Figure \ref{chart02}.

It can be observed from Figure \ref{chart02} (a) to (d) that, using the calculation method of Eq. (\ref{eq02}), the value of cutoff distance $d_{c}$ has a great influence on the local density of the data points.
When $d_{c}$ is equal to 0.57 (1.0\%), the densities of most data points are 1.0 while delta distances are in the range of 0.0 to 10.0.
There is only one point with high density (2.0) and high delta distance (20.1), which is the candidate cluster center.
With the change of the value of $d_{c}$, the distributions of decision points in Figure \ref{chart02} (b), (c), and (d) are changed evidently.
In contrast, it can be found from Figure \ref{chart02} (e) to (h) that, using GKF of Eq. (\ref{eq04}), the value of cutoff distance has minimal effect on the local density of the data points.
When the value of cutoff distance $d_{c}$ increases from 0.57 (1.0\%) to 0.88 (4.0\%), the density distribution of the data points is basically stable.
There are three decision points with high density and high delta distance (shown as the red squares) in the four cases of (e) to (h), which are the candidate cluster centers.
At the same time, there are four decision points with low density and high delta distance (shown as the black triangles) in the four cases, which refer the outliers.
Moreover, the three candidate cluster centers and four outliers refer to the same data points in the four cases.
Among them, four outliers are the 1-st (AFP = 1.06 and CEA = 10.70), 2-nd (AFP = 2.32 and CEA = 13.50), 12-th (AFP = 7.38 and CEA = 10.35), and 13-th (AFP = 8.00 and CEA = 8.10) of data points in Table \ref{table04}.
Three candidate cluster centers are the 18-th (AFP = 4.54 and CEA = 26.03), 22-th (AFP = 4.91 and CEA = 27.98), and 26-th (AFP = 5.30 and CEA = 27.62) of data points, which refer different degrees of disease symptom of liver cancer.
Therefore, we can carefully draw the conclusion that, the DPCA algorithm achieves high robustness by using GKF and obtain accurate disease-symptom clusters.

\subsubsection{Quality Evaluation of Treatment Recommendation}
\label{section_QualityEvaluation}

\begin{figure}[!ht]
\setlength{\abovecaptionskip}{0pt}
\setlength{\belowcaptionskip}{0pt}
\centering
\includegraphics[width=3.0in]{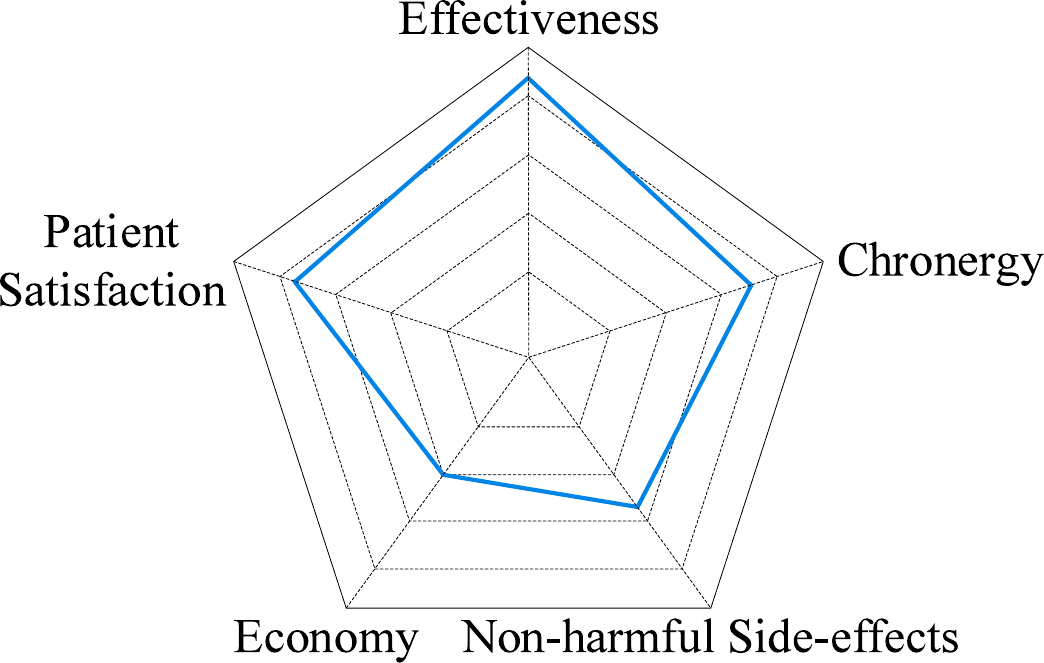}
\caption{Radar graph of the quality of treatment schemes}
\label{chart03}
\end{figure}

To evaluate the quality of the treatment recommendation that DDTRS provided, an evaluation is performed via feedback from medical doctors.
There are five indicators available on the application interface to evaluate the quality of the treatment scheme: (a) effectiveness, (b) chronergy, (c) non-harmful side-effects, (d) economy, and (e) patient satisfaction.
The quality of a treatment scheme is defined in Eq. (\ref{eq16}):

\begin{equation}
\label{eq16}
Q = (Eft, ~Chr, ~NSE, ~Eco, ~PS),
\end{equation}
where $Eft$ is the effectiveness of the treatment scheme, $Chr$ is the chronergy, $NSE$ is the non-harmful side-effects, $Eco$ is the economy, and $PS$ is the patient satisfaction.
The scoring range of each indicator is (1 $\sim$ 5).
The quality $Q$ of the treatment scheme is drawn as a radar graph, as shown in Figure \ref{chart03}.

The radar graph of $Q$ is a pentagon, in which the distance from the center to each vertex is equal to 5.
It is easy to obtain that each side length is approximately 5.88 and the area of the pentagon is $S_{pentagon} = \frac{(5.88)^{2}}{4} \sqrt{25+10\sqrt{5}} \approx 59.44$.
Hence, the value of $Q$ is within the range of ($1 \sim 59.44$).
According to the radar graph, the value of $Q$ is the area defined by the five indicators (Eq. (\ref{eq17})):

\begin{equation}
\label{eq17}
\begin{aligned}
Q = & \frac{Eft \times Chr + Chr \times NSE + NSE \times Eco}{2} \times \sin {72^{o}}\\
    & + \frac{Eco \times PS + PS \times Eft}{2} \times \sin {72^{o}}.
\end{aligned}
\end{equation}

In Eq. (\ref{eq17}), the value of each indicator is computed with the scores submitted by doctors and their weights.
Taking the effectiveness indicator $Eft$ of a treatment plan as an example, assume that $N$ scores are fed back from doctors.
Then, $Eft$ is defined in Eq. (\ref{eq18}):

\begin{equation}
\label{eq18}
Eft =\sum_{i=1}^{N}{\left( eft_{i} \times w_{i}\right)},
\end{equation}
where $eft_{i}$ is the score evaluated by the $i$-th doctor and $w_{i}$ is his weight, with a value in the range of (0 $\sim$ 1).

\begin{figure}[!ht]
\setlength{\abovecaptionskip}{0pt}
\setlength{\belowcaptionskip}{0pt}
\centering
\includegraphics[width=3.5in]{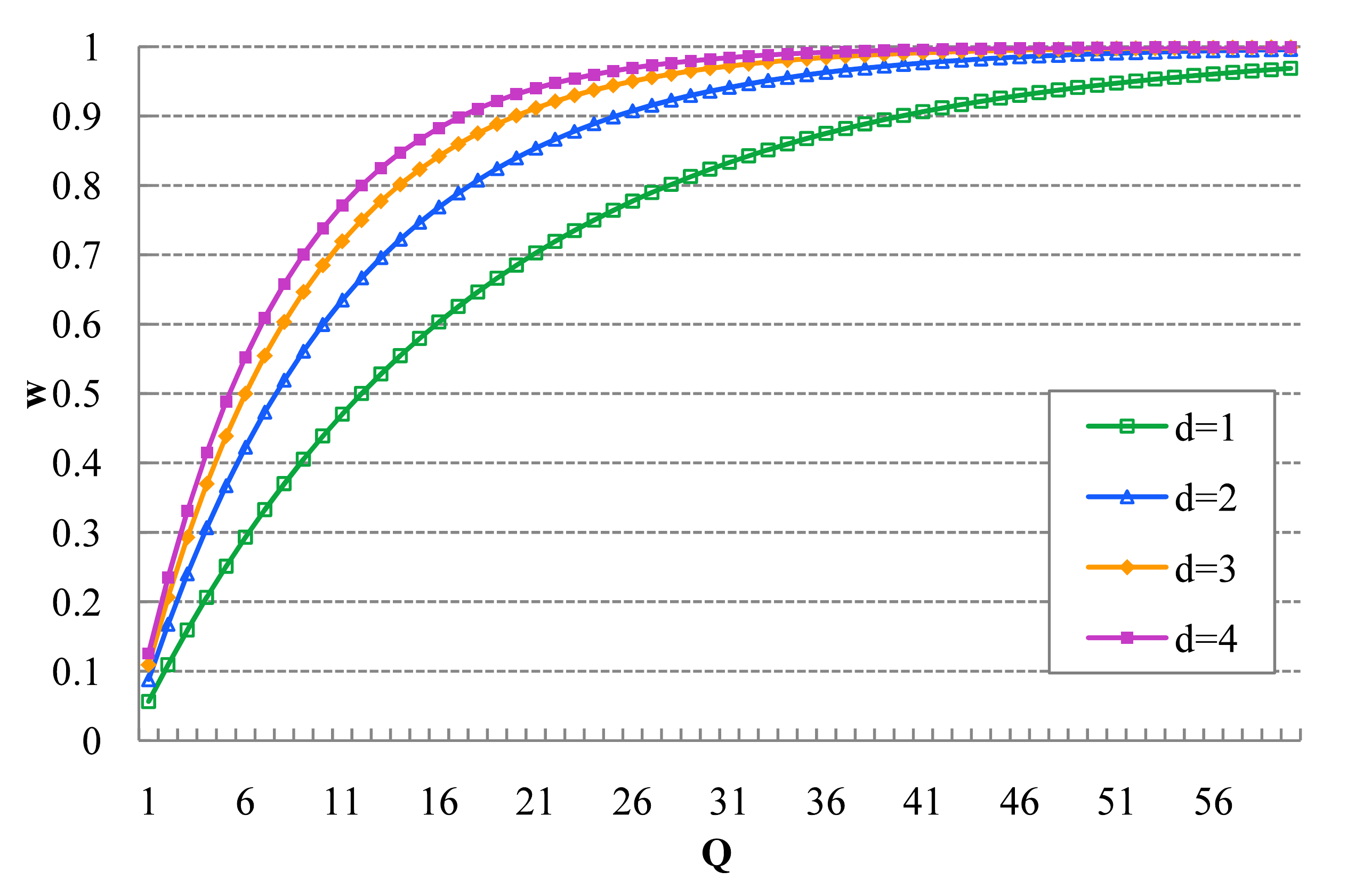}
\caption{Weights of doctors with different professional titles}
\label{chart04}
\end{figure}

The weight of a doctor is defined based on the professional authority of the doctor in his field of expertise.
Namely, it's value is calculated by combining his professional title and the average quality of his historical treatment schemes.
Assuming that $M$ treatment schemes presented by the $i$-th doctor are utilized in the recommendation, let $\bar{Q_{i}} = \frac{1}{M}\sum_{j=1}^{M}{Q_{j}}$ be the average quality of all treatment schemes of the doctor.
The weight value of the $i$-th doctor {is} defined in Eq. (\ref{eq19}):

\begin{equation}
\label{eq19}
\begin{aligned}
w_{i} &= -(d_{i}+1)^{\frac{1}{12}\bar{Q_{i}}}+1,\\
\end{aligned}
\end{equation}
where $d_{i}$ is the professional degree of the doctor, the value of which is in (1, 2, 3, 4), corresponding to different professional titles, such as primary (physician / resident), intermediate (doctor), deputy senior (deputy chief physician), and senior (chief physician).
If there is no treatment scheme belonging to this doctor, we set $\bar{Q_{i}} = 1$.
The weight distribution of doctors with different professional titles is shown in Figure \ref{chart04}.

\begin{figure}[!ht]
\setlength{\abovecaptionskip}{0pt}
\setlength{\belowcaptionskip}{0pt}
\centering
\includegraphics[width=3.0in]{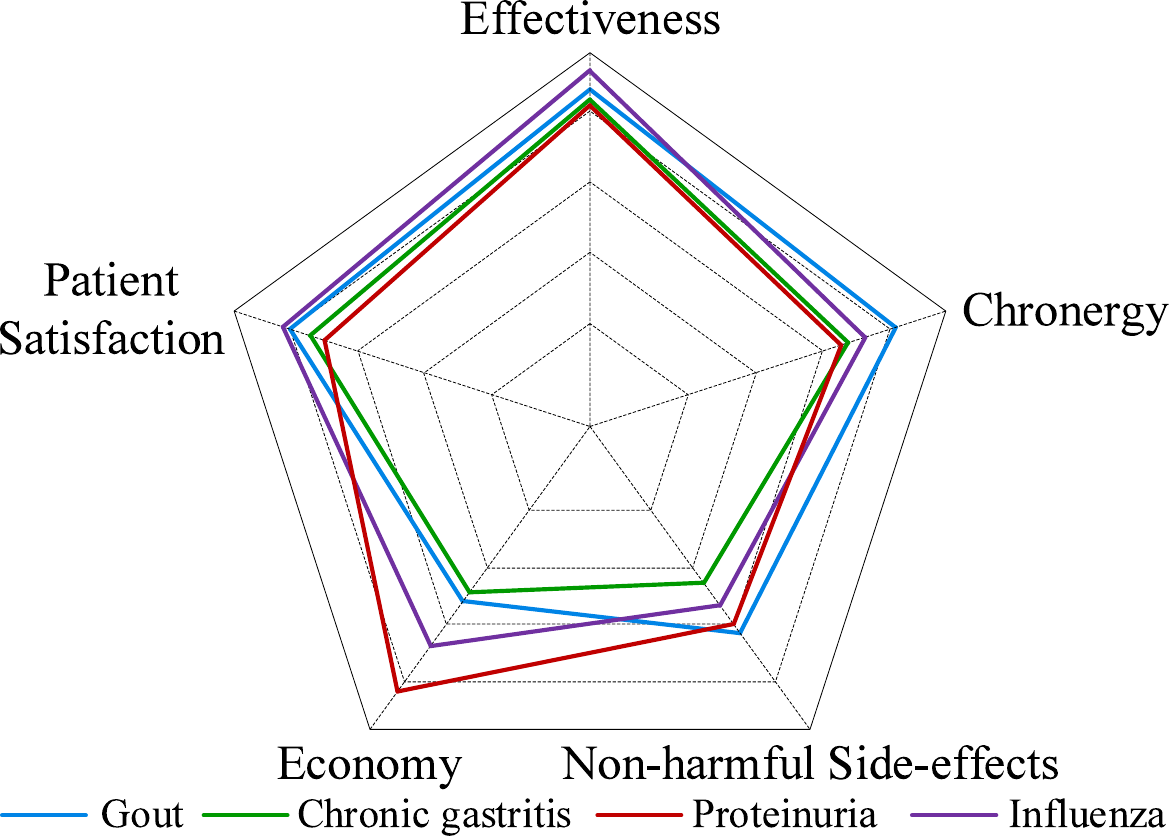}
\caption{Quality evaluation of the treatment recommendation}
\label{chart05}
\end{figure}

Four treatment schemes are considered in this experiment to evaluate to quality of the recommendation module.
For each treatment-scheme case, more than 10,000 feedbacks over a period of more than six months are gathered.
The results of the quality evaluation of treatment recommendations are shown in Figure \ref{chart05}, and the detailed description is provided in Table \ref{table05}.

\begin{table}[!ht]
\setlength{\abovecaptionskip}{0pt}
\setlength{\belowcaptionskip}{0pt}
\renewcommand{\arraystretch}{1.3}
\caption{Detailed description of the quality evaluation}
\centering
\label{table05}
\begin{tabular}{clcccccc}
\hline
No.&	Treatment Scheme& $Eft$ & $Chr$ &  $NSE$& $Eco$& $PS$& $Q$ \\
\hline
1 & Gout & 4.32 & 4.19 & 4.20 & 2.86 & 4.02 & 36.41\\
2 & Chronic gastritis & 4.11 & 3.39 & 4.02 & 4.14 & 3.62 & 35.22 \\
3 & Proteinuria & 4.15 & 3.39 & 3.32 & 2.61 & 3.82 & 28.44 \\
4 & Influenza & 4.73 & 3.71 & 3.89 & 3.33 & 4.15 & 37.27 \\
\hline
\end{tabular}
\end{table}

As shown in Figure \ref{chart05} and Table \ref{table05}, the four treatment schemes reach overall quality scores of 28.44 - 37.27, with an average of 34.34.
Compared with the other four indicators, the effectiveness achieves the highest overall score, with high values in the range of 4.11 - 4.73 (82.20\% - 94.60\%) and 4.33 (86.60\%) on average.
Indicators of non-harmful side-effects and patient satisfaction are ranked second and third, with average values of 3.90 (78.00\%) and 3.86 (77.20\%), respectively.
However, the economic indicator is ranked the lowest, with a value of 3.24 (64.80\%) on average.
In general, a high-quality treatment recommendation means that the treatment has high effectiveness, chronergy, and few side effects.
The experimental results demonstrate that the system can provide high-quality treatment recommendations and achieve the desired design goals.

\subsection{Application of Disease Treatment Recommendation}
\label{section5.3_CaseAnalysis}

\subsubsection{Case Analysis of Liver Cancer}

To estimate the validity and reliability of DDTRS, an application of disease-symptom clustering is considered and treatment recommendation is performed.
DDTRS is deployed at our cooperating hospitals in China.
A tumor-marker inspection task is performed on a male patient volunteer of age 50.
The inspection report of the tumor marker stored in HDFS is described as follows.

\textit{
``Tumor marker inspection:: specimen : serum, patient gender : female, age : 50, CEA : 5.56, AFP : 121000, CA724 : 0.69, CA125 : 28.29, CA153 : 11.86,CA199 : 0.60, NSE : 11.30, GLO : 29.4, HCG-$\beta$ : 4.17, GGT : 12.0, ...''
}

After clustering the submitted inspection report, a disease-symptom cluster named $Liver ~Cancer$ is obtained from similar historical inspection data.
Compared with the clustering results, diseases are usually inaccurately diagnosed by inexperienced medical doctors according to their own experience.
The cluster center of the $Liver ~Cancer$ disease is detected in the form of mean symptom values, which are characterized as follows.

\textit{
``Tumor marker inspection:: specimen : serum, CEA : 6.62, AFP : 113052, CA724 : 1.37, CA125 : 30.13, CA153 : 10.21, CA199 : 3.80, NSE : 10.78, GLO : 31.7, HCG-$\beta$ : 5.28, GGT : 23.0, ...''
}

Depending on the disease-symptom cluster $Liver ~Cancer$, related diagnosis schemes and treatment plans are recommended:

\textit{
``The symptoms reflect the liver cancer.
Liver cancer is the malignant tumor that occurs in the liver, which including a primary liver cancer and a metastatic liver cancer.
When AFP is more than 25 ng/ml and CEA is more than 5 ng/ml, it is often suggested as a tumor, which is common in liver cancer, colorectal cancer, and breast cancer.''
}

The inspection report, disease-symptom clustering, and the treatment recommendation for $Liver ~Cancer$ are presented in Figure \ref{fig09}.
The recommended treatment scheme for $Liver ~Cancer$ is surgical resection, which is the best choice.
It can lead to healing by complete removal of the tumor tissue.
The medication scheme of $Liver ~Cancer$ is suggested in Table \ref{table06}.

\begin{table}[!ht]
\centering
\renewcommand{\arraystretch}{1.3}
\setlength{\abovecaptionskip}{0pt}
\setlength{\belowcaptionskip}{0pt}
\caption{Medication plan of the $Liver ~Cancer$ disease}
\label{table06}
\begin{tabular}{clccc}
\hline
No. & Drug name & Dosage & Route & Times per day\\
\hline
1 & Gefitinib & 250 mg & 250 mg & 1  \\
2 & Erlotinib & 150 mg & 150 mg & 3 \\
3 & Icotinib  & 125 mg & 125 mg & 3  \\
4 & Rh2       & 100 mg & 100 ml & 1\\
\hline
\end{tabular}
\end{table}

\subsubsection{Case Analysis of Anti-Cardiolipin Antibody Syndrome}

Another case of disease-symptom clustering and treatment recommendation is considered to verify the effectiveness of the proposed system.
In the first place, a blood serum inspection task is performed on a female volunteer of age 33, who has been pregnant for  two months.
The inspection report of the blood serum for the volunteer is stored on HDFS and is characterized as follows.

\textit{
``Blood serum:: specimen : serum, patient gender : female, age : 33, ACA-IgM : 13.6, ACA-IgG : 6.4, $\beta2$-GP1 : 93.2, ANA : 5.11,
AcBA : 0, dsDNA : 13.08, AnuA : 0, AJo-1 : 0, ArPO : 0, ARNP/Sm : 0, ARo-52 : 1, ASc1-70 : 0, ASm : 0, ASS-A : 1, ASS-B : 0, AenA : 1.''
}

After submitting the report to the disease-symptom clustering module, a disease-symptom cluster named $Anti$-$Cardiolipin$ $Antibody$ $Syndrome$ $(ACAS)$ is derived.
Compared with the clustering result, inaccurate disease classes are usually diagnosed by inexperienced doctors based on their own experience, such as a $Systemic$ $Lupus$ $Erythematosus$ $(SLE)$ or a $Rheumatoid$ $Arthritis$ $(RA)$.
According to the disease-symptom cluster $ACAS$, the related diagnosis schemes and treatment schemes are recommended.
The recommended diagnosis scheme is described as follows.

\textit{
``ACAS is a general term for a group of clinical signs caused by an anti-cardiolipin antibody.
Anti-cardiolipin antibody is a type of autoantibody that the target antigen of it is taken on the platelets and the heart phospholipids, which exists on the endothelial cell membrane with negatively charged.
ACAS often appears in SLE and other autoimmune diseases.
There are close relationships between this antibody and the diseases, such as thrombosis, throm bocytopenia, spontaneous abortion, and intrauterine fetal death.''
}

The recommended treatment plan for the $ACAS$ disease is to inject heparin sodium and take medication at the same time; the medication plan for $ACAS$ is shown in Table \ref{table07}.

\begin{table}[!ht]
\centering
\renewcommand{\arraystretch}{1.3}
\setlength{\abovecaptionskip}{0pt}
\setlength{\belowcaptionskip}{0pt}
\caption{Medication plan for the $ACAS$ disease}
\label{table07}
\begin{tabular}{clccc}
\hline
No. & ~~~~~~~~~~~~~~Drug name & Dosage & Route & Times per day\\
\hline
1 & Aspirin & 100 mg & 100 mg & 1  \\
2 & Meprednisone & 4 mg & 4 mg & 1 \\
3 & Hydroxychloroquine Sulfate Tablets & 100 mg & 200 mg & 2   \\
4 & Low-molecular-weight Heparin Sodium Injection & 0.4 ml 5000 IU & 0.4 ml & 1\\
\hline
\end{tabular}
\end{table}

\subsection{Performance Evaluation}
\label{section5.4_PerformanceEvaluation}

\subsubsection{Performance of Disease-Symptom Clustering}
\begin{figure}[!ht]
\setlength{\abovecaptionskip}{0pt}
\setlength{\belowcaptionskip}{0pt}
\centering
\includegraphics[width=3.5in]{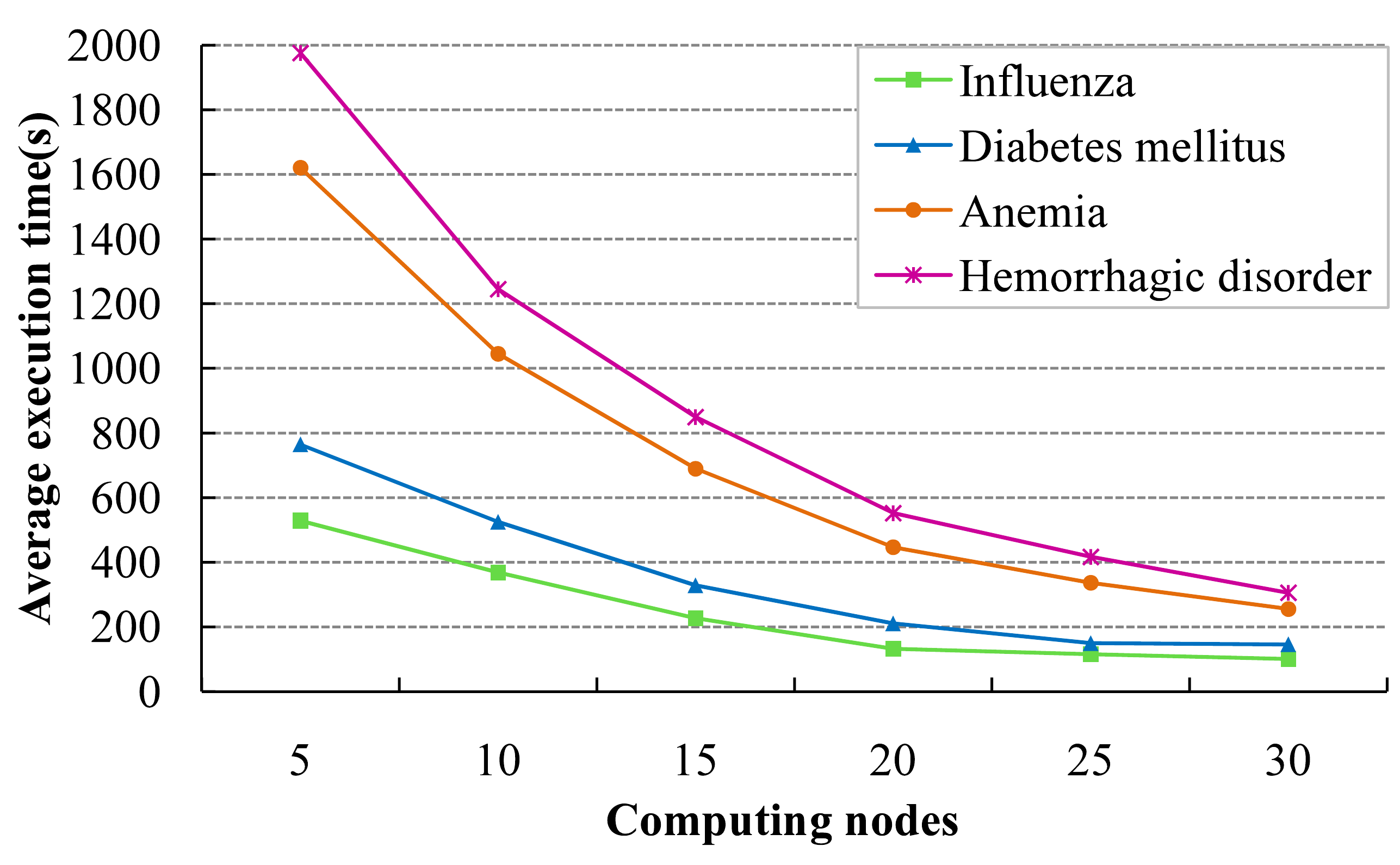}
\caption{Performance evaluation of the disease-symptom clustering}
\label{chart06}
\end{figure}

The performance of the disease-symptom clustering module of DDTRS is assessed.
Inspection reports of four types of diseases are clustered, namely, influenza, diabetes mellitus, anemia, and hemorrhagic disorder.
Their data sizes are 5GB, 10GB, 15GB, and 20GB, respectively.
The experiments on each group of inspection datasets are executed on the Spark cluster.
On the Spark cluster, the number of computing nodes increases from 5 to 30 in each case of the experiments.
By observing the average execution time of the disease-symptom clustering process, performance across different cases is compared and analyzed.
The results of the performance evaluation experiments are presented in Figure \ref{chart06}.

In Figure \ref{chart06}, when the number of the cluster nodes is set to 5, due to the different sizes of the inspection data, the execution time of the clustering process for each case is different evidently.
For example, in the case of hemorrhagic disorder, owing to the data size of 20GB, the execution time is longer than those of all the other cases, peaking at 1976.56 s.
In contrast, the execution time is 1621.25 s in the case of anemia (15GB), 764.37 s in the case of diabetes mellitus (10GB), and 528.52 s in the case of influenza (5GB).
As the number of cluster nodes increases from 5 to 30, taking advantage of data-parallel and task-parallel schemes of parallel computing, the performance of the disease-symptom clustering process is improved distinctly.
For example, the average execution time decreases from 528.52 s to 103.68 s in the case of influenza and decreases from 764.37 s to 145.52 s for diabetes mellitus.
Obviously, the advantage of parallel optimization is greater in cases of large-scale datasets than in cases of small-scale datasets.
The benefit is more noticeable when the dataset size increases or the number of slave nodes increases to a finite extent.

\subsubsection{Performance of Disease-treatment Association Analysis and Recommendation}

\begin{figure}[!ht]
\setlength{\abovecaptionskip}{0pt}
\setlength{\belowcaptionskip}{0pt}
\centering
\includegraphics[width=3.5in]{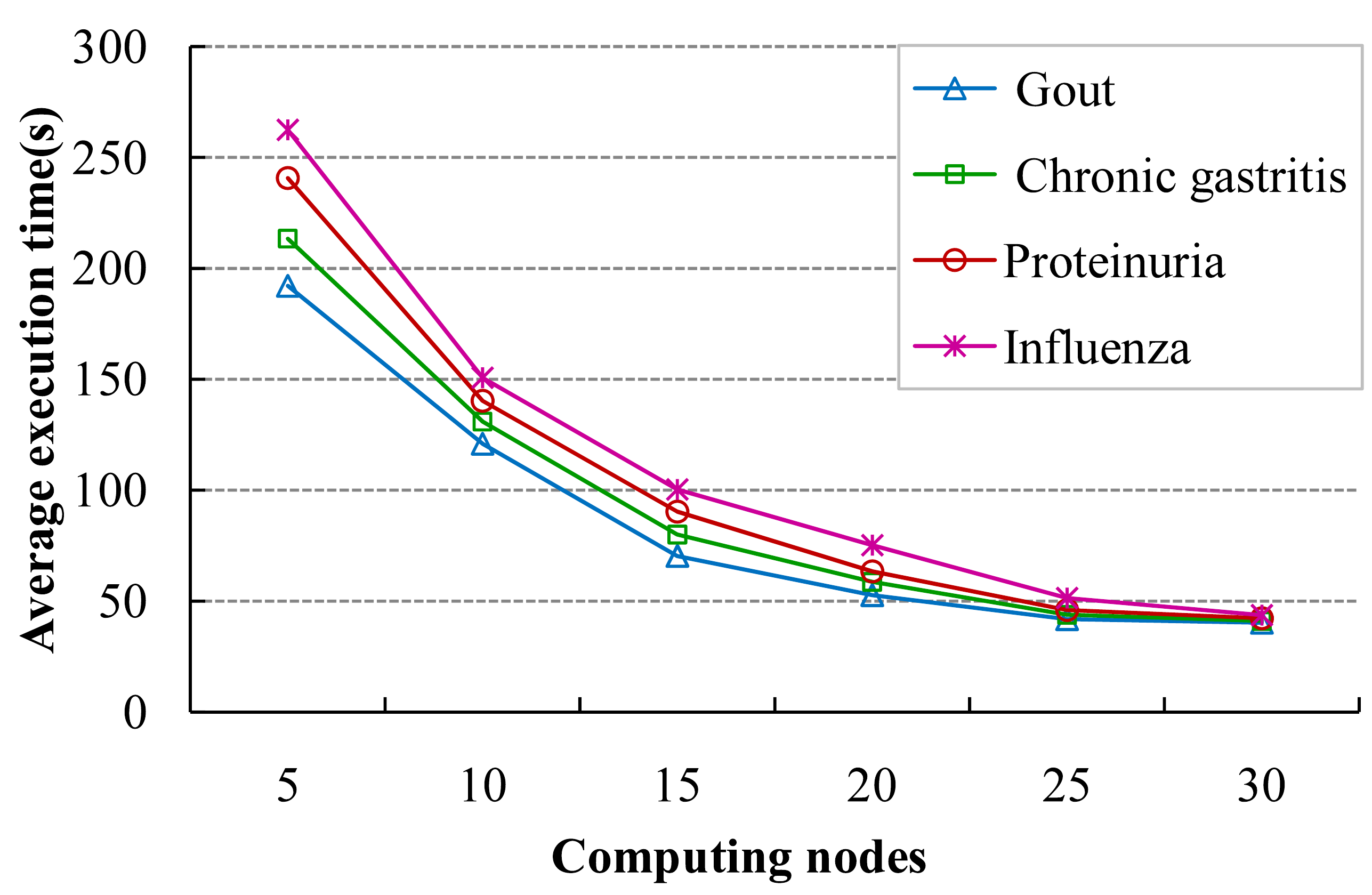}
\caption{Performance evaluation of the disease-treatment association analysis process}
\label{chart07}
\end{figure}

To evaluate the performance of the disease treatment association analysis and recommendation module, four groups of datasets of disease diagnosis and treatment are analyzed on the Spark cluster under different scales.
The number of computing nodes in each experiment increases from 5 to 30.
The average execution times of the association analysis process in different cases are illustrated in Figure \ref{chart07}.

When there are five computing nodes on the Spark cluster, the average execution times of the association analysis process are 262.45 s for influenza, 240.71 s for proteinuria, 213.36 s for chronic gastritis, and 192.12 s for gout.
As the cluster scale increases from 5 to 30, the average execution times of the association analysis process decrease by ratios of 6.02, 5.70, 5.20, and 4.78, respectively.
Almost-linear speedup is achieved.
The average execution time in the four cases is approximately 41.76 s.
Benefitting from the high computing power of cloud computing and parallel computing, the performance of DDTRS is improved significantly.

\section{Conclusions}
This paper presented a disease diagnosis and treatment recommendation system to recommend medical treatments based on the given inspection reports of patients.
The proposed system first clusters the inspection reports to obtain disease symptoms as clustering centers.
Then, it identifies strong links between diseases and treatments by applying an association analysis algorithm.
The paper further provided qualitative analysis of the proposed system by demonstrating how a patient and a doctor obtain treatment recommendations based on the patient's inspection reports.
Experimental and application results indicated that the system effectively achieves the required goals and provides high-quality recommendations with low latency response.

In our future work, the effectiveness of the disease diagnosis and treatment schemes will be evaluated by tracking the changes in the inspection parameters of each patient.
In addition, appropriate and accurate recommendations will be offered based on superior weighted diagnosis and treatment schemes.
Moreover, security perspectives of the system in an actual application environment will be addressed.

\section*{Acknowledgment}
We would like to thank the editors and the anonymous reviewers for their constructive comments and criticisms.
This research was partially funded by the Key Program of the National Natural Science Foundation of China (Grant No. 61432005),
the National Outstanding Youth Science Program of National Natural Science Foundation of China (Grant No. 61625202),
the National Natural Science Foundation of China (Grant Nos. 61672221),
the China Scholarships Council (Grant Nos. 201706130080),
and the Hunan Provincial Innovation Foundation For Postgraduate (Grant No. CX2017B099).

\bibliography{RMERbib}

\end{document}